# Graphical Fermat's Principle and Triangle-Free Graph Estimation

Junwei Lu,* Han Liu†

**Abstract**

We consider the problem of estimating undirected triangle-free graphs of high dimensional distributions. Triangle-free graphs form a rich graph family which allows arbitrary loopy structures but 3-cliques. For inferential tractability, we propose a graphical Fermat's principle to regularize the distribution family. Such principle enforces the existence of a distribution-dependent pseudo-metric such that any two nodes have a smaller distance than that of two other nodes who have a geodesic path include these two nodes. Guided by this principle, we show that a greedy strategy is able to recover the true graph. The resulting algorithm only requires a pairwise distance matrix as input and is computationally even more efficient than calculating the minimum spanning tree. We consider graph estimation problems under different settings, including discrete and nonparametric distribution families. Thorough numerical results are provided to illustrate the usefulness of the proposed method.

## 1 Introduction

Graphical model provides a powerful tool to explore complex distributions. Let $\boldsymbol{X} = (X_1, \ldots, X_d)^T$ be a $d$-dimensional random vector with distribution $P_{\boldsymbol{X}}$. We denote the graph of $P_{\boldsymbol{X}}$ to be $G = (V, E)$, where the vertex set $V$ corresponds to the variables $X_1, \ldots, X_d$ and the edge set $E$ characterizes the conditional independence relationships between these variables. Specifically, two nodes $X_i$ and $X_j$ are not connected if and only if they are conditionally independent given the other variables. To utilize graphical models, a fundamental problem is to estimate the graph based on observational data.

In a graph estimation problem, we observe $n$ samples from $P_{\boldsymbol{X}}$ and aim to infer the structure of the graph $G$. Many existing graph estimation methods involve two steps: (1) graph metric estimation: estimating a pairwise "distance" matrix defined by some (pseudo) metric and (2) structure learning: applying a structure learning algorithm based on the estimated pairwise "distance" matrix to recover the graph structure. For example, for the Gaussian graphical model, the graph metric on an edge

*Department of Operations Research and Financial Engineering, Princeton University, Princeton, NJ 08544, USA; Email: `junweil@princeton.edu`

†Department of Operations Research and Financial Engineering, Princeton University, Princeton, NJ 08544, USA; Email: `hanliu@princeton.edu`

$(i, j)$ is defined as $d_{ij} = |\boldsymbol{\Omega}_{ij}|$, where $\boldsymbol{\Omega}$ is the inverse covariance matrix. The metric estimation step estimates the inverse covariance matrix (Meinshausen and Bühlmann, 2006; Banerjee et al., 2008; Rothman et al., 2008; Friedman et al., 2008; d'Aspremont et al., 2008; Yuan, 2010; Cai et al., 2011; Liu and Wang, 2012; Ren et al., 2014) and the structure learning step adds an edge $(i, j)$ if and only if $d_{ij} \neq 0$. For the Ising model, the graph metric for each edge is the absolute value of the interaction parameter on the edge. The metric estimation step can be conducted by sparse logistic regression (Ravikumar et al., 2010) and the structure learning step is determined by the sparsity pattern of the estimated coefficients from the sparse logistic regression. A more general framework on generalized linear graphical models has been proposed by Yang et al. (2012). They assume the nodewise-conditional distributions follow generalized linear models and the values of the parameter vector can be interpreted as graph metrics. Other more complex graphical models include the nonparanormal graphical model (Liu et al., 2012b; Xue and Zou, 2012) and transelliptical graphical model (Liu et al., 2012a), which consider the inverse of Kendall's tau matrix as the graph metric and estimate the graph by its sparsity. Voorman et al. (2014) assume the conditional means of variables are additive models and propose a semiparametric method to estimate the graph.

The above graph estimation methods are either parametric or semiparametric. To further relax the distributional assumption, we need to enforce more constraints on the estimated graphs. One popular constraint is to enforce the estimated graph to be a tree. Under this structural regularization, Mossel (2007) considers the information distance $d_{ij} = -\log|\det P_{\boldsymbol{X}_{i,j}}|$ as graph metric for discrete distributions. Here $P_{\boldsymbol{X}_{i,j}}$ is the joint probability matrix for $\boldsymbol{X}_i, \boldsymbol{X}_j$ and it can be estimated based on the empirical distribution. In the structure learning step, methods solving minimum spanning tree (MST) like Kruskal's algorithm (Kruskal, 1956) or Chow-Liu algorithm (Chow and Liu, 1968) can be applied with the edge weights determined by the estimated graph metrics. Another example is nonparametric forest graphical models. Under the same tree structural regularization, Liu et al. (2011) use mutual information as the graph metric and suggest to apply the maximum spanning tree algorithm as the second step for graph recovery. Assuming the existence of possible latent variables, Song et al. (2011) consider a nonparametric graph metric based on the pseudo-determinant of the covariance operator. They apply the spectral methods to estimate the graph metric and the recursive grouping algorithm (Choi et al., 2011) to estimate the graph structure. To relax the restrictive tree structure, Anandkumar and Valluvan (2013) consider a locally tree-like graphical model. Such model allows loopy structures whose girth is long enough such that the neighborhood of any node is a tree. The information distance is shown to be a proper graph metric if the model satisfies the correlation decay condition (Georgii, 2011; Mézard and Montanari, 2009; Weitz, 2005). For graph estimation, they propose to use minimum spanning tree algorithm to learn the tree neighborhoods of each node and merge these local trees together. In addition to the tree or locally tree-like graph structure assumption, Loh and Wainwright (2013) and Loh and Bühlmann (2014) propose to learn the graph by estimating the inverse generalized covariance matrix for certain discrete distribution families.

In this paper, we propose a new graph estimation method which allows more general graph structures and can handle both parametric and nonparametric graphical models under a unified regularization framework called graphical Fermat's principle. The graphical Fermat's principle regularizes the graphical models by assuming the existence of some pseudo-metric defined as a functional of the joint distribution $P_{\boldsymbol{X}}$ such that the metric between any two nodes $i, j$ is larger than the metric between nodes $i', j'$ if both $i', j'$ lies in the geodesic connecting $i, j$. The corresponding metric is called Fermat metric. The graphical Fermat's principle is a kind of variational principle for pairwise conditional dependency on the graphical models. In particular, it characterizes the phenomenon that nodes with stronger dependency have shorter paths connecting them. Our graph learning method estimates the pairwise Fermat metrics in the metric estimation step. For the structure learning step, we propose a minimum triangle-free graph (MTG) estimation algorithm to recover the graph. Our method has three advantages over the existing methods. First, our method allows arbitrary graph structure without 3-cliques. Compared to existing algorithms which require the girth of a graph is long enough, we can handle more complicated graph structures as long as its girth is larger than 3. Second, the graphical Fermat's principle holds for a large family of parametric and nonparametric graphical models. We show that the information distance $d_{ij} = -\log|\det P_{\boldsymbol{X}_{i,j}}|$ for the locally tree-like graphical model, the negative mutual information $-I(X_i, X_j)$ for the Gaussian tree model, and the nonparametric tree metric defined for latent tree model are all Fermat metrics. Third, our structure learning method MTG is computationally more efficient than MST. Within each iteration, MST checks whether a new edge forms a cycle but MTG only needs to check whether it forms a triangle. Moreover, let $d_{ij}$ be the Fermat metric between nodes $i, j$. If there exists a sequence of $r_n$ such that the minimum metric gap between edges has $\min_{(i,j),(i',j')\in E} |d_{ij} - d_{i'j'}| \gtrsim (\log d)/r_n$ and the Fermat metric estimator $\widehat{d}_{ij}$ has an exponential concentration $\mathbb{P}(|\widehat{d}_{ij} - d_{ij}| > \epsilon) < C_1 \exp(-C_2 r_n \epsilon^2)$ for all $i, j = 1, \ldots, d$, we show that the graph can be consistently recovered using MTG. As two applications, we consider both discrete and nonparametric graphical models. For nonparametric models using the negative mutual information as the Fermat metric, we also propose a new robust rank-based mutual information estimator, which is applicable to density function with arbitrary support.

## 1.1 Paper Organization

The rest of the paper is organized as follows. In Section 2, we introduce the graphical Fermat's principle. The relationship between the Fermat metric and other existing graph metrics is also discussed. In Sections 3, we propose a minimum triangle-free graph (MTG) estimation algorithm and prove its consistency. In Section 4, we apply the proposed algorithm to both discrete and nonparametric graphical models and prove their theoretical properties. Section 5 illustrates the numerical performance of our method.

## 2 Preliminaries and Notations

Let $\boldsymbol{X} := (X_1, \ldots, X_d)^T$ be a $d$-dimensional random vector. If $\boldsymbol{X}$ is Markov to the graph $G = (V, E)$, its joint density $p_G(\boldsymbol{x})$ bears the factorization

$$p_G(\boldsymbol{x}) = \exp\left(\sum_{c \in \mathcal{C}} \theta_c(\boldsymbol{x}_c) - A(\theta)\right), \tag{2.1}$$

where $\mathcal{C}$ is the set of cliques in $G$, $\boldsymbol{x}_c$ is a vector indexed by the clique $c$, and $A(\theta)$ is a probability density normalizer. The factorization in (2.1) reveals the topological structure of the graphical model. For example, the Gaussian Markov random field is a special family of graphical model with the density factorization

$$p_G(\boldsymbol{x}) = \exp\left(-\sum_{(i,j)\in E} \boldsymbol{\Omega}_{ij}(x_i-\mu_i)(x_j-\mu_j) - \frac{1}{2}\sum_{k=1}^d \boldsymbol{\Omega}_{kk}(x_k-\mu_k)^2 - \frac{1}{2}\log[(2\pi)^d \det(\boldsymbol{\Sigma})]\right), \quad (2.2)$$

where $\det(\cdot)$ is the determinant, $\boldsymbol{\Sigma}$ is the covariance matrix and $\boldsymbol{\Omega} = \boldsymbol{\Sigma}^{-1}$. The factorization in (2.2) implies that $(i,j) \in E$ if and only if $\boldsymbol{\Omega}_{ij} \neq 0$. The Ising model characterizing binary values $\{+1,-1\}^d$ follows the probability mass function

$$P_G(\boldsymbol{x}) = \exp\left(\sum_{(i,j)\in E} \theta_{ij} x_i x_j + \sum_{k=1}^d \phi_k x_k - A(\boldsymbol{\theta})\right), \quad (2.3)$$

for all $\boldsymbol{x} \in \{+1,-1\}^d$. The graphical structure can be recovered by the support of the interaction parameters $\{\theta_{ij}\}_{i,j=1}^d$. A forest model is a nonparametric graphical model whose graph is a forest $F = (V, E)$ and its density $p_F(\cdot)$ can be written as

$$p_F(\boldsymbol{x}) = \prod_{(i,j)\in E} \frac{p(x_i, x_j)}{p(x_i)p(x_j)} \prod_{k=1}^d p(x_k). \quad (2.4)$$

Here $p(x_i, x_j)$ is the bivariate marginal density of $X_i$ and $X_j$, and $p(x_k)$ is the univariate marginal density of $X_k$. All the above models have some distribution functionals between every two nodes $i \neq j \in \{1, \ldots, d\}$: $\boldsymbol{\Omega}_{ij}$ for Gaussian graphical model, $\theta_{ij}$ for Ising model, and the mutual information

$$I(X_i; X_j) := \iint p(x_i, x_j) \log\left(\frac{p(x_i, x_j)}{p(x_i)p(x_j)}\right) dx_i\, dx_j \quad (2.5)$$

for forest graphical model. They all encode the underlying graph of the model. To formalize these relationships, we propose the graphical Fermat's principle, which provides a framework for a large variety of graph estimation problems.

## 2.1 Graphical Fermat's Principle and Fermat Metric

In this subsection, we present the definition of the graphical Fermat's principle and Fermat metric. More properties of the Fermat metric are provided later. We also discuss the relations between the Fermat metric and other existing graph metrics in the literature. Before rigorously define the graphical Fermat's principle, we explain its motivation by reviewing the forest density estimator (FDE) proposed by Liu et al. (2011). For each pair of nodes $i, j \in V$, FDE associates $(i, j)$ with the mutual information $I(X_i; X_j)$ as the weight and applies the maximum spanning tree algorithm to learn the graph. This is equivalent to set the graph metric to be $d_{ij} = -I(X_i, X_j)$ for all $i, j$ and find a minimum spanning tree (MST) based on $\{d_{ij}\}_{i,j\in V}$. The reason why MST works for FDE is that more "important" edges with stronger dependency have smaller graph metric as $d_{ij} = -I(X_i; X_j)$. Therefore, we tend to add edges with smaller graph metric as early as possible and MST matches this intuition. In the following, we denote the graph metric between any two vertices $i, j \in V$ as $d_{ij} \in \mathbb{R} \cup \{+\infty\}$[1]. The graphical Fermat's principle specifies that a good graph metric $\{d_{ij}\}_{i,j\in V}$ for structure learning should have the property that smaller $d_{ij}$ implies that $(i, j)$ is more possible to be a true edge. The following definition formalizes this intuition.

---

[1] One thing to note is that our definition of graph metric is quite general. It does not need to be nonnegative or do not have to satisfy triangle inequality.

**Definition 2.1** (Graphical Fermat's Principle). Suppose a random vector $\boldsymbol{X} \in \mathbb{R}^d$ has a graphical density $p_G(\boldsymbol{x})$ and $G = (V, E)$ is the associated graph. We say the path connecting $i, j$:

$$(i, u_1) \to (u_1, u_2) \to (u_2, u_3) \to \ldots \to (u_{m-2}, u_{m-1}) \to (u_{m-1}, j)$$

is of length $m$ if $i \neq u_1 \neq \ldots \neq u_{m-1} \neq j$. Let $\text{Path}_{ij}(m)$ be the set of edges consisting of the paths connecting $i, j$ such that

$$\text{Path}_{i,j}(m) = \{e \in E \mid \text{edge } e \text{ is on a path connecting } i, j \text{ with length equal or smaller than } m\}.$$

The graph geodesic between $i, j$ is the path with shortest length connecting $i, j$ and we denote the length of the graph geodesic between $i, j$ to be $d(i, j)$. The graph geodesic is denoted as $\text{Path}(i, j) := \text{Path}_{i,j}(d(i, j))$. A graph metric $\{d_{ij}\}_{i,j \in V}$ is a Fermat metric if and only if for any $i, j, u, v \in V$, the graph geodesics $\text{Path}(i, j)$ and $\text{Path}(u, v)$ connecting $u, v$ and $i, j$ satisfy

$$\text{Path}(u, v) \subsetneq \text{Path}(i, j) \quad \text{implies} \quad d_{uv} < d_{ij}. \tag{2.6}$$

If there exists a graph metric $\{d_{ij}\}_{i,j \in V}$ satisfying (2.6), we say that the graphical model $p_G(\boldsymbol{x})$ follows the graphical Fermat's principle.

Graphical Fermat's principle is an analogue to Fermat's principle in physics describing the behavior of light: it tends to take the path which can be traversed in the least time. In graphical model, the graphical Fermat's principle assumes that the nodes connected by shorter paths tend to have stronger dependency. Indeed we can derive the following variational proposition of Fermat metric from (2.6).

**Proposition 2.2.** (Variational Property of the Fermat Metric) If the graph metric $\{d_{ij}\}_{i,j \in V}$ is a Fermat metric, we have for any $i, j \in V$ that $(i, j) \notin E$,

$$d_{ij} > \max_{(u,v) \in \text{Path}(i,j)} d_{uv}. \tag{2.7}$$

*Proof.* For any $(u, v) \in \text{Path}(i, j)$ such that $(u, v) \neq (i, j)$, we have $\text{Path}(u, v) \subsetneq \text{Path}(i, j)$, since for any path connecting $u, v$, we can construct a path connecting $i, j$ by combining the $\text{Path}(u, v)$ and the rest of $\text{Path}(i, j)$. According to (2.6), $d_{ij} > d_{uv}$. Since $(u, v)$ is arbitrary, (2.7) is proved. □

If we interpret the Fermat metric $\{d_{ij}\}_{i,j \in V}$ as a pseudo distance, the variational property of the Fermat metric in Proposition 2.2 tells us that the distance between two nodes is larger than the geodesic path connecting them. Another interesting property of the Fermat metric is that (2.6) is invariant under strictly monotone transformation. In particular, as (2.6) only relies on the inequality, a Fermat metric remains to be a Fermat metric after any strictly monotone transformation. This implies that if there exists one Fermat metric on the graphical model, we can derive infinite number of Fermat metrics by applying strictly monotone transformations. In the following, we compare the Fermat metric with the tree metric in Song et al. (2011), and give several concrete examples of the Fermat metrics.

**Definition 2.3** (Tree Metric (Song et al., 2011)). Let $T = (V, E)$ be a tree, the graph metric $\{d_{ij}\}_{i,j \in V}$ is a tree metric (also called information distance by Erdős et al. (1999) and Choi et al. (2011)) if for every node $i, j \in V$, we have

$$d_{ij} = \sum_{(u,v) \in \text{Path}(i,j)} d_{uv}, \tag{2.8}$$

where $\text{Path}(i, j)$ is the set of edges on the unique path connecting $i, j$.

Comparing Definition 2.1 and Definition 2.3, since there is only one path connecting two nodes on a tree, we can immediately derive that a tree metric $\{d_{ij}\}_{i,j \in V}$ must be a Fermat metric. On the other hand, a Fermat metric is not necessarily a tree metric. First, Fermat metric can be defined on any kind of graphs while a tree metric can only be defined on a tree graphical model. Second, if we compare (2.7) and (2.8), the tree metric requires additivity of metrics on a path while (2.7) is much weaker. Furthermore, if we apply certain monotone transform on a tree metric, it will remain to be a Fermat metric even if it could be no longer a tree metric. We list several examples of tree metrics as follows. According to the discussion above, they are also examples of Fermat's metrics.

**Example 2.4** (Discrete Distribution(Lake, 1994)). Suppose $\boldsymbol{X} = (X_1, \ldots, X_d)^T$ follows the discrete tree distribution, we define $d_{ij}$ for every $i, j$ that

$$d_{ij} = -\log \det(P_{ij}) + \frac{1}{2}\log \det(P_{ii}) + \frac{1}{2}\log \det(P_{jj}), \tag{2.9}$$

where $P_{ij}$ is the discrete joint probability matrix for $X_i, X_j$ and $P_{ii}, P_{jj}$ are diagonal matrices with marginal probabilities of $X_i$ and $X_j$ on their diagonals.

**Example 2.5** (Gaussian Markov Random Field). Let $\boldsymbol{X} = (X_1, \ldots, X_d)^T \sim p_T(\boldsymbol{x})$ be a Gaussian distribution that is Markov to a tree $T$. We define the graph metric as the negative mutual information $-I(X_i; X_j) = -1/2 \cdot \log(1 - \rho_{ij}^2)$, where $\rho_{ij}$ is the correlation between $X_i$ and $X_j$.

**Example 2.6** (Nonparametric Tree Graphical Model). A random vecotr $(X_1, \ldots, X_d)^T \sim p_T(\boldsymbol{x})$ follows a nonparametric tree distribution. Let $\mathcal{H}$ be a reproducing kernel Hilbert space (RKHS) and the corresponding kernel is $K(x, y)$. By the Mercer theorem, there exists a feature map $\phi(x) \in \mathcal{H}$ such that $K(x, x') := \langle \phi(x), \phi(x') \rangle_{\mathcal{H}}$. We define the covariance operator $\mathcal{C}_{ij} := \mathbb{E}_{X_i X_j}[\phi(X_i) \otimes \phi(X_j)]$ for $i, j \in V$. Here $\otimes$ is a tensor product. For any $f, g \in \mathcal{H}$, $f \otimes g : \mathcal{H} \mapsto \mathcal{H}$ is an operator such that $f \otimes g(h) = \langle g, h \rangle_{\mathcal{H}} f$ for any $h \in \mathcal{H}$. The pseudo-determinant of the operator $\mathcal{C}$ is the product of non-zero singular values of $\mathcal{C}$ and we denote $|\mathcal{C}|_\star = \prod_i \sigma(\mathcal{C})$. For any $i, j \in V$, the metric is defined as

$$d_{ij} = -\frac{1}{2}\log|\mathcal{C}_{ij}\mathcal{C}_{ij}^T|_\star + \frac{1}{4}\log|\mathcal{C}_{ii}\mathcal{C}_{ii}^T|_\star + \frac{1}{4}\log|\mathcal{C}_{jj}\mathcal{C}_{jj}^T|_\star. \tag{2.10}$$

**Proposition 2.7.** The graph metrics $d_{ij}$ in Example 2.4, Example 2.5, Example 2.6 and their strict monotone transformations are all Fermat metrics.

*Proof.* Erdős et al. (1999) prove that the graph metric in (2.9) for discrete tree model and $d_{ij} = -\log|\rho_{ij}|$ for Gaussian tree model are both tree metrics and therefore are Fermat metric. Similarly, Song et al. (2011) also prove that the pseduo-determinant is a tree metric. Since we have $-I(X_i; X_j) = -1/2 \cdot \log(1 - \rho_{ij}^2) = -1/2 \cdot \log(1 - \exp(-2d_{ij}))$ for every $i, j \in V$ being a strictly monotone transformation of $d_{ij}$. Therefore $-I(X_i; X_j)$ is a Fermat metric for Gaussian graphical models Markov to trees. □

### 2.1.1 Comparison to Correlation Decay

In addition to the graphical Fermat's principle, correlation decay is another regularity assumption that has been popularly used for inferring loopy graphs. It is first raised in statistical physics (Georgii, 2011; Mézard and Montanari, 2009; Weitz, 2005) and applied in statistics by Montanari and Pereira (2009) and Anandkumar et al. (2012) for the recovery of Ising models. Anandkumar and Valluvan (2013) applies it to the structure estimation of discrete latent graphical model by a local Chow-Liu grouping algorithm (LocalCLGrouping). In order to recover a graph with cycles, LocalCLGrouping constructs local subtrees of the graph by Chow-Liu grouping method and pieces them together. The algorithm is valid when the graph is locally tree-like, which is guaranteed by correlation decay.

In this section, we discuss the relationship between the graphical Fermat's principle and the correlation decay property. In particular, we show that the correlation decay property secures the existence of a valid Fermat metric. This allows us to construct concrete examples of Fermat metrics on graphical models even with loopy structures.

Before introducing the definition of correlation decay, we start with some notations. Let $P_{\mathbf{X}|G}$ be the discrete distribution that is Markov to $G$ and $P_{\mathbf{X}_A|F}$ be the marginal distribution on the edge set $A \subset E$ with the potentials on the edges $E \setminus F$ being zero. We denote the girth of $G$ by $g$. We assume the random variable $X_i$ takes values on a finite set $\mathcal{X}$. Recall that the graph distance between two nodes is $d(i,j)$ and we can therefore induce the distance between two sets of nodes $E_1, E_2 \subset E$ and $d(E_1, E_2) = \min_{i \in E_1, j \in E_2} d(i,j)$. We denote $B_l(i) := \{j \in V : \text{dist}(i,j) \leq l\}$ and $\partial B_l(i) := \{j \in V : d(i,j) = l\}$ is the nodes with graph distances equals to $l$. Let $F_l(i;G) := G(B_l(i))$ be the subgraph induced by the nodes in $F_l(i;G)$. Denote $\|P - Q\|_1$ be the $\ell_1$ norm of two discrete distribution such that $\|P - Q\|_1 := \sum_{s \in \mathcal{S}} |P(s) - Q(s)|$, where $\mathcal{S}$ is the state space.

**Definition 2.8** (Correlation Decay (Anandkumar and Valluvan, 2013)). The discrete graphical model $P_{\mathbf{X}_A|G}$ that is Markov to $G = (V, E)$ is said to have correlation decay if for all $l \in \mathbb{N}, i \in V$ and $A \subset B_l(i)$,
$$\|P_{\mathbf{X}_A|G} - P_{\mathbf{X}_A|F_l(i;G)}\|_1 \leq \zeta(d(A, \partial B_l(i))), \tag{2.11}$$
where $\zeta$ is a nonincreasing rate function characterizes the decay of correlation.

The correlation decay in (2.11) implies that there is no long-range correlation in the graphical model such that it behaves locally like a tree. The distance considered in Anandkumar and Valluvan (2013) is the information distance $d_{ij} = -\log|\det(P_{X_{i,j}})|$ for all $i, j \in V$. As the graphical model is locally tree-like, the local tree metric is defined as $d(i,j;\text{tree}) := -\log|\det P_{\mathbf{X}_{i,j}|\text{tree}(i,j)}|$, where $\text{tree}(i,j) := G\big(B_l(i) \cup B_l(j)\big)$.

**Assumption 2.9.** The following assumptions are needed by Anandkumar and Valluvan (2013) for the consistency of graph recovery of LocalCLGrouping.

1. The marginal distributions of local tree $P_{\mathbf{X}_{i,j}|\text{tree}(i,j)}$ are nonsingular for all $i, j \in V$ and
$$0 < d_{\min} \leq d(i,j;\text{tree}) \leq d_{\max} < \infty, \quad \eta := \frac{d_{\max}}{d_{\min}}.$$

2. The nonincreasing rate function $\zeta(\cdot)$ satisfies
$$0 \leq \zeta\left(\frac{g}{2} - l - 1\right) < \frac{\nu}{|\mathcal{X}|^2}, \text{ where}$$
$$\nu = \min\left(d_{\min}, 0.5 e^{-l d_{\min}}(e^{d_{\min}} - 1), e^{-0.5 d_{\max}(l+2)}, \left(\frac{g}{4} - l\right) d_{\min}\right).$$

**Theorem 2.10.** The discrete graphical model $P_{\boldsymbol{X}_A|G}$ satisfies the correlation decay in (2.11). We define the gap between pairwise information distances $d_{ij} = -\log|\det(P_{\boldsymbol{X}_{i,j}})|$ as $\delta_{\max} = \max_{m=1,\ldots,d^2-1}|d_{(m)} - d_{(m+1)}|$, where $d_{(m)}$ is the $m$-th largest information distance, and $d'_{\max}(l) := ld_{\max} - \log(1 - e^{ld_{\max}}\zeta(g/2-l-1))$. Under Assumption 2.9, if $\delta_{\max} \geq 2|\mathcal{X}|e^{d'_{\max}(l)}\zeta(g/2 - l - 1)$, then $\{d_{ij}\}_{i,j \in V}$ is a Fermat metric.

The proof of the above theorem is deferred to Appendix A. A sufficient condition of correlation decay for Ising models in (2.3) can be explicitly established by the restrictions on the edge potentials of Ising models. See more details in Appendix A.1.

## 3 Method of Graph Estimation

In this section, we introduce the minimum triangle-free graph algorithm (MTG), which reconstructs the underlying graph with loops but no 3-cliques. Suppose that we already have an estimator for the Fermat metric $\{\widehat{d}_{ij}\}_{i,j \in V}$. We take the $\{\widehat{d}_{ij}\}_{i,j \in V}$ as the input of our algorithm. Concrete examples of estimating $\{d_{ij}\}_{i,j \in V}$ are shown in Section 4.

### 3.1 Graph Estimation Algorithm

The minimum triangle-free graph algorithm is similar to MST. If we denote $\mathcal{T}(V)$ to be the set of all trees with vertex set $V$ and $\mathsf{MST}(V; \mathbf{d})$ be the minimum spanning tree with vertex set $V$ and edge weights $\{d_{ij}\}_{i,j \in V}$, where

$$\mathsf{MST}(V; d_{ij}) := \arg\min_{T \in \mathcal{T}(V)} \sum_{(i,j) \in T} d_{ij}.$$

The output $\mathsf{MST}(V; d_{ij})$ is called a Chow-Liu tree (Since Chow and Liu (1968) applied the Kruskal's algorithm Kruskal (1956) to estimate a tree density). Within each iteration, the algorithm greedily adds the edge with smallest weight among the set of edges not yet visited if no cycle is generated. It is described in detail in Algorithm 1.

---
**Algorithm 1** Minimum Spanning Tree (Chow and Liu, 1968)

**Input:** Graph metric estimator $\widehat{\mathbf{d}} = \{\widehat{d}_{ij}\}_{i,j \in V}$
Initialize $E_0^{\mathrm{t}} = \varnothing$
**for** $k = 1, \ldots, d^2$ **do**
 $(i_k, j_k) \leftarrow \arg\min_{i,j}\{\widehat{d}_{ij}\}_{(i,j) \notin E_{k-1}}$
 $E_k \leftarrow E_{k-1} \cup (i_k, j_k)$
 $E_k^{\mathrm{t}} \leftarrow E_{k-1}^{\mathrm{t}} \cup \{(i_k, j_k)\}$ if $E_{k-1}^{\mathrm{t}} \cup \{(i_k, j_k)\}$ does not contain a cycle
**end for**
**Output:** The minimum spanning tree $\widehat{F}_{d^2} = (V, E_{d^2}^{\mathrm{t}})$.

---

Similar to MST, the intuition behind the proposed MTG algorithm is that the connected edges in the true graph generally have smaller graph metric, and an edge should not be added if it violates the graphical Fermat's principle. The algorithm of MTG is simple: First, we sort all the edges according to their weights (from large to small); Second, we greedily add in these edges according to the sorted order. Whenever a new edge is added in, we need to make sure that no triangle is formed. Otherwise, this edge should not be added in. The detailed procedure of MST is described in Algorithm 2. The only difference between MST and MTG is the mechanism of deleting edges. MST deletes edges when detecting a new cycle and MTG deletes edges when detecting a new triangle. This makes MTG more efficient than MST in computation since triangles are much easier to be detected. The other difference is that MST detects cycles in $E_k^t$ defined in Algorithm 1. However, MTG deletes $(i_k, j_k)$ when a new triangle formed in $E_k$ instead of $E_k^g$ as indicated in Algorithm 2.

---

**Algorithm 2** Minimun Triangle-free Graph

---

**Input:** Graph metric estimator $\widehat{\mathbf{d}} = \{\widehat{d}_{ij}\}_{i,j \in V}$
Initialize $E_0 = E_0^g = \varnothing$
**for** $k = 1, \ldots, d^2$ **do**
  $(i_k, j_k) \leftarrow \arg\min_{i,j}\{\widehat{d}_{ij}\}_{(i,j) \notin E_{k-1}}$
  $E_k \leftarrow E_{k-1} \cup (i_k, j_k)$
  **if** $E_{k-1} \cup (i_k, j_k)$ does not form a new triangle **then**
    $E_k^g \leftarrow E_{k-1}^g \cup (i_k, j_k)$
  **else**
    $E_k^g \leftarrow E_{k-1}^g$
  **end if**
**end for**
**Output:** The minimum triangle-free graph $\widehat{G}_{d^2} = (V, E_{d^2}^g)$.

---

Algorithm 1 deletes the $k$-th edge $(i_k, j_k)$ when $E_{k-1}^t \cup (i_k, j_k)$ contains a cycle. Algorithm 2 deletes $(i_k, j_k)$ when $E_{k-1} \cup (i_k, j_k)$ contains a triangle. Since $E_{k-1}^t$ and $E_{k-1}$ are different, without any assumption, $E_k^g$ does not necessarily contain $E_k^t$. However, the following theorem shows that if $\{d_{ij}\}_{i,j \in V}$ is a Fermat metric, $E_k^t$ is always a subset of $E_k^g$.

**Theorem 3.1.** Let $\widetilde{G}_k = (V, E_k)$, $\widehat{F}_k = (V, E_k^t)$ and $\widehat{G}_k = (V, E_k^g)$ be the graphs in the $k$-th iteration of Algorithm 1 and Algorithm 2 with the input $\widehat{\mathbf{d}} = \{\widehat{d}_{ij}\}_{i,j \in V}$, then $\widehat{F}_k$ is the subgraph of $\widehat{G}_k$ for every $k = 0, 1, \ldots, d-1$. The minimum spanning tree $\widehat{F}_{d-1}$ is therefore called the forest skeleton of the minimum triangle-free graph $\widehat{G}_{d-1}$. In particular, we have the following filtration chart:

$$
\begin{array}{ccccccccc}
\text{Greedy:} & \varnothing & \subset & \widetilde{G}_1 & \subset & \widetilde{G}_2 & \subset & \cdots & \subset & \widetilde{G}_{d-1} \\
& & & \cup & & \cup & & & & \cup \\
\text{Triangle-free:} & \varnothing & \subset & \widehat{G}_1 & \subset & \widehat{G}_2 & \subset & \cdots & \subset & \widehat{G}_{d-1} \\
& & & \cup & & \cup & & & & \cup \\
\text{Chow-Liu:} & \varnothing & \subset & \widehat{F}_1 & \subset & \widehat{F}_2 & \subset & \cdots & \subset & \widehat{F}_{d-1}.
\end{array}
\quad (3.1)
$$

**Remark 3.2.** The proof is shown in Appendix B. From Theorem 3.1, it is easy to see why the graphical Fermat's principle can only be applied to learning triangle-free graphs. Considering a 3-clique graph, suppose two edges have been added. We cannot decide whether the third edge with the smallest metric should be added or not, since both cases do not violate the graphical Fermat's principle. Therefore the Fermat metric cannot identify triangles in the graphical models.

### 3.2 Graph Estimation Consistency

In this section, we establish the graph estimation consistency of Algorithm 2 based on the graphical Fermat's principle. We first consider the case that the input is the true Fermat metric. In the following of paper, we assume that the Fermat metric always exists on the given graphical models and is known.

**Assumption 3.3** (Fermat Metric). For the graphical model $p_G(\boldsymbol{x})$, there exists a graph metric $\{d_{ij}\}_{i,j=1}^d$ being a Fermat metric defined in Defintion 2.1.

Under this assumption, we obtain the exact recovery result of Algorithm 2 as follows. The proof is deferred to the Appendix C.

**Lemma 3.4.** Let the associated graph $G = (V, E)$ of a graphical model $p_G(\boldsymbol{x})$ be a triangle-free graph. If the input $\mathbf{d} = \{d_{ij}\}_{i,j\in V}$ of Algorithm 2 is the population Fermat metric, Algorithm 2 can exactly recover the true graph $G$.

The output of Algorithm 2 only depends on the relative order of the Fermat metric on each edge. If a gap between two pairs of metrics $d_{jk}$ and $d_{i'j'}$ is too small, it is hard to guarantee that the estimated Fermat metric on each edge follows the same order as the truth. However, for graph estimation, it is unnecessary to recover the relative order exactly. Since some changes of the relative order do not influence the output graph, we just need to ensure that those relative orders which change the output are recovered precisely. Define the crucial set $\mathcal{C} \subset E \times E$ such that $(e, e') \in \mathcal{C}$ if and only if flipping the relative order of $d_e$ and $d_{e'}$ will change the output graph of Algorithm 2. The following assumption gives the condition which guarantees the graph estimation consistency of MTG.

**Assumption 3.5.** Let $\mathcal{C}$ be the crucial set. There exists a positive sequence of $r_n \to \infty$ as $n \to \infty$ such that the gap between the Fermat metric on $\mathcal{C}$ satisfies

$$\min_{(e,e')\in\mathcal{C}} |d_e - d_{e'}| \geq L_n, \text{ where } L_n = \Omega\left(\sqrt{\frac{\log d + \log n}{r_n}}\right), \tag{3.2}$$

where $L_n = \Omega(a_n)$ denotes that there exists a constant $C$ such that $L_n \geq C a_n$ for sufficiently large $n$.

Liu et al. (2011) define a similar crucial set on MTG and require a similar assumption as Assumption 3.5. The $r_n$ in (3.2) is a generic notation related to the estimation rate of the metric estimator. It will be specified in the next section for concrete examples. The following theorem characterizes how $r_n$ plays the role in terms of graph recovery.

**Theorem 3.6 (Graph Recovery Consistency).** Suppose there exists a positive sequence of $r_n$ satisfying Assumption 3.5. Under Assumption 3.3, if the Fermat metric estimator $\widehat{\mathbf{d}} = \{\widehat{d}_{ij}\}_{i,j\in V}$ has the following exponential concentration inequality related to $r_n$ for all $\epsilon \geq L_n$ in (3.2)

$$\mathbb{P}\left(|\widehat{d}_{ij} - d_{ij}| > \epsilon\right) \leq C_1 \exp\left(-C_2 r_n \epsilon^2\right), \tag{3.3}$$

where the constants $C_1, C_2$ are independent to $i, j$. The estimated $\widehat{G}$ from Algorithm 2 has $\lim_{n\to\infty} \mathbb{P}\left(\widehat{G} \neq G\right) = 0$.

**Remark 3.7.** We prove the results in Appendix D. In the theorem, the sequence $\{1/\sqrt{r_n}\}_{n=1}^\infty$ represents the rate of convergence for the Fermat metric and it varies with different graphical models and diverse metric estimators. Assumption 3.5 and $L_n$ in (3.2) quantify how big the gap should be. In the next section, we will provide explicit rates of $r_n$ and $L_n$ for concrete graphical models.

# 4 Graph Estimation Based on Fermat Metrics

In this section, we apply MTG to learn both the discrete and nonparametric graphical models. The exponential concentration inequalities of Fermat metric estimators similar to (3.3) are proved such that graph estimation consistency can be guaranteed according to Theorem 3.6.

## 4.1 Graph Estimation for Discrete Graphical Models

Suppose $\boldsymbol{X} = (X_1, \ldots, X_d)^T$ is a discrete random vector whose output set $\mathcal{A}$ is finite. As shown in Example 2.4, we can define an information distance $d_{ij} = -\log\det(P_{ij}) + \frac{1}{2}\log\det(P_{ii}) + \frac{1}{2}\log\det(P_{jj})$ for any $i, j \in V$. It has been proved in Proposition 2.7 that if the graph $G$ of $\boldsymbol{X}$ is a tree, the information distance $\{d_{ij}\}_{i,j\in V}$ is a Fermat metric.

Let $\widehat{P}_{ii}, \widehat{P}_{jj}$ and $\widehat{P}_{ij}$ be the empirical marginal probability matrices and the information distance estimator is $\widehat{d}_{ij} = -\log\det(\widehat{P}_{ij}) + \frac{1}{2}\log\det(\widehat{P}_{ii}) + \frac{1}{2}\log\det(\widehat{P}_{jj})$. We have the following lemma on its estimation rate, whose proof is deferred to Appendix E.

**Lemma 4.1.** Suppose the marginal bivariate joint stochastic matrix $P_{ij}$ is nonsingular for every $i, j \in V$. Furthermore, there exists a constant $p_0 > 0$ such that $p_0^{-1} \leq |P_{ij}| \leq p_0$ for any $i, j$. We have, for all $\epsilon > 4p_0^{-s/2}$ where $s = |\mathcal{A}|$,

$$\mathbb{P}\left(|\exp(-\widehat{d}_{ij}) - \exp(-d_{ij})| > \epsilon\right) < 2^{s+2}\exp\left(-\frac{p_0^{4s}n\epsilon^2}{32s^2}\right). \tag{4.1}$$

Define the new graph metric $d'_{ij} = -\exp(-d_{ij})$ for any $i, j \in V$. Since $d'_{ij}$ is a monotone transform of $d_{ij}$, it is still a Fermat metric. We now apply the MTG to the new Fermat metric estimator $\widehat{d}'_{ij} = -\exp(-\widehat{d}_{ij})$ for all $i, j \in V$ and obtain the estimated graph $\widehat{G}$. According to Theorem 3.6, we have the following graph estimation consistency result for discrete graphical model.

**Theorem 4.2.** Suppose the information distance $\{d_{ij}\}_{i,j\in V}$ in (2.9) is a Fermat metric for the discrete graphical model and there exists a constant $p_0 > 0$ such that $p_0^{-1} \leq |P_{ij}| \leq p_0$ for any $i, j$. The gap between different Fermat metrics satisfies

$$\min_{(e,e')\in\mathcal{C}} |\exp(-d_e) - \exp(-d_{e'})| \geq L_n, \tag{4.2}$$

where $L_n = \Omega(\sqrt{(\log d + \log n)/n})$. We obtain the $\widehat{G}$ from Algorithm 2 by inputting $\widehat{d}'_{ij} = -\exp(-\widehat{d}_{ij})$. If $L_n < 4p_0^{-s/2}$, we have $\lim_{n\to\infty}\mathbb{P}\left(\widehat{G} \neq G\right) = 0$.

*Proof.* As the new graph metric $\{d'_{ij}\}_{i,j\in V}$ is still a Fermat metric and it satisfies Assumption 3.5 for $r_n = n$ due to (4.2), Lemma 4.1 gives us the concentration inequality of the metric estimator for all $\epsilon > 4p_0^{-s/2} > L_n$ required in (3.3). By Theorem 3.6, we prove the graph estimation consistency. □

## 4.2 Graph Estimation for Nonparametric Graphical Models

Let $\boldsymbol{x}_1, \ldots, \boldsymbol{x}_n$ be $n$ data points generated from a $d$-dimensional random vector $\boldsymbol{X} := (X_1, \ldots, X_d)^T$. We denote $\boldsymbol{x}_i := (x_{i1}, \ldots, x_{in})^T$. Suppose $\boldsymbol{X}$ is Markov to the graph $G = (V, E)$ and the joint density is $p_G(\boldsymbol{x})$. In this section, we consider nonparametric graphical models with the negative mutual information as their Fermat metric. In addition, we propose a robust mutual information estimator based on rank statistics and show its statistical properties.

### 4.2.1 Kernel Mutual Information Estimator

A naive way to estimate the mutual information is to apply its definition in (2.5) and construct a plug-in estimator

$$I(\widehat{p}_h) := \iint \widehat{p}_h(x_1, x_2) \log\left(\frac{\widehat{p}_h(x_1, x_2)}{\widehat{p}_h(x_1)\widehat{p}_h(x_2)}\right) dx_1\, dx_2, \tag{4.3}$$

where $\widehat{p}_h(x_i, x_j)$ and $\widehat{p}_h(x_k)$ are the bivariate and univariate distribution estimators either through the empirical distribution or the kernel density estimator. Specifically, the kernel density estimator of the bivariate and univariate density functions are

$$\widehat{p}_h(x_i, x_j) := \frac{1}{nh^2}\sum_{k=1}^{n} K\left(\frac{x_i - x_{ik}}{h}\right)K\left(\frac{x_j - x_{jk}}{h}\right),\ \widehat{p}_h(x_u) := \frac{1}{nh}\sum_{k=1}^{n} K\left(\frac{x_u - x_{uk}}{h}\right), \tag{4.4}$$

for every $i, j, u = 1, \ldots, d$. Because the estimator in (4.3) needs double integration on $\mathbb{R}^2$, existing methods (Liu et al., 2011, 2012c) generally assume the density function that its support must be on a unit square. However in practical applications, a large family of random variables have unbounded supports. In order to bridge the gap between the real world applications and theoretical analysis, we propose a robust estimator applicable for unbounded random variables. Our estimator is grounded on a key observation that the mutual information is equivalent to a negative copula entropy. In particular, the copula of $(X_i, X_j)$ is defined as the joint cumulative distribution function:

$$C(u_i, u_j) = \mathbb{P}(U_i \leq u_i, U_j \leq u_j) = \mathbb{P}(X_i \leq (F^{(i)})^{-1}(u_i), X_j \leq (F^{(j)})^{-1}(u_j)), \tag{4.5}$$

where $F^{(\ell)}(x_\ell) = \mathbb{P}(X_\ell \leq x_\ell)$ is the cumulative distribution function of $X_\ell$, and $U_\ell = F^{(\ell)}(X_\ell)$. The corresponding copula density is $c(u_i, u_j) = \partial^2 C/\partial u_i \partial u_j$. It is known that $(U_i, U_j)$ has uniform marginal distributions (Sklar, 1959). The information of the marginal distribution of $(X_i, X_j)$ is eliminated in $(U_i, U_j)$ and the copula density characterizes the dependency between $X_i$ and $X_j$. Moreover, there is a direct connection between the copula entropy and mutual information:

$$I(X_i; X_j) = -H_c(X_i, X_j) = \iint c(u_i, u_j) \log c(u_i, u_j) du_i du_j. \tag{4.6}$$

This is a key property for our robust mutual information estimator. From (4.6), we derive a novel approach to estimate the mutual information by estimating the copula density through the kernel density estimation based on the rank statistic $(\widehat{U}_i, \widehat{U}_j) := (F_n^{(i)}(X_i), F_n^{(j)}(X_j))$ where $F_n^{(k)}$ is the empirical marginal distribution function of $F^{(k)}$. Consider a bivariate random vector $(X_i, X_j)$ with joint marginal density $p(x_i, x_j)$. We denote the marginal distribution function of $X_i, X_j$ as $F^{(i)}(x), F^{(j)}(x)$. $F_n^{(i)}$ and $F_n^{(j)}$ are the corresponding empirical cumulative distribution functions of the marginals of $p(x_i, x_j)$ where

$$F_n^{(k)}(x) = \frac{1}{n}\sum_{k=1}^{n}\mathbb{1}\{X_k \leq x\},\ \text{for } k = 1, \ldots, d.$$

Denote $U_k = F^{(k)}(X_k), \widehat{U}_k = F_n^{(k)}(X_k)$ and the samples for $i = 1, \ldots, n$ are defined as

$$\begin{aligned} \boldsymbol{u}_i &:= (u_{i1}, \ldots, u_{in})^T = (F^{(i)}(x_{i1}), \ldots, F^{(i)}(x_{in}))^T, \\ \widehat{\boldsymbol{u}}_i &:= (\widehat{u}_{i1}, \ldots, \widehat{u}_{in})^T = (F_n^{(i)}(x_{i1}), \ldots, F_n^{(i)}(x_{in}))^T. \end{aligned}$$

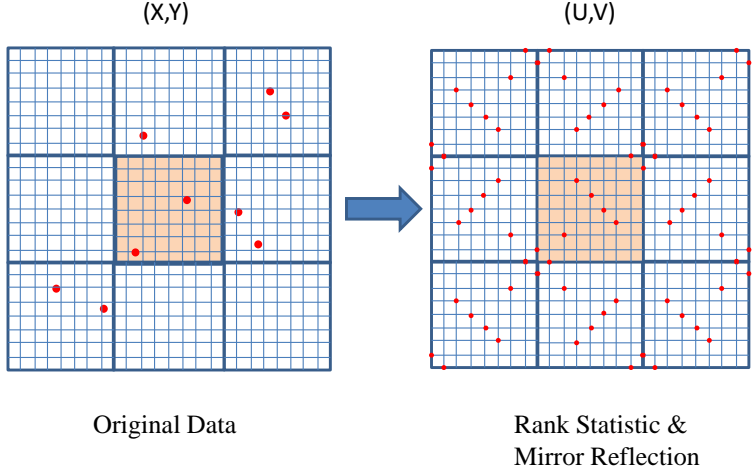

Figure 1: The procedure of converting the data $\{\boldsymbol{x}_i\}_{i=1}^n$ to the rank statistics $\{\widehat{\boldsymbol{u}}_i\}_{i=1}^n$ and operating the mirror reflection.

Let $c(u_i, u_j)$ be the density of copula distribution $C(u_i, u_j)$ defined in (4.5). According to (4.6), in order to estimate the mutual information, it suffices to estimate the copula entropy.

We apply the kernel density estimation to infer the copula density. To reduce the boundary bias, we use the "mirror reflection" kernel density estimator introduced in Gijbels and Mielniczuk (1990):

$$\widetilde{c}_h(u_i, u_j) := \frac{1}{nh^2} \sum_{k=1}^n \sum_{\ell=1}^9 \left\{ K\left(\frac{u_i - \widehat{u}_{ik}^{(\ell)}}{h}\right) K\left(\frac{u_j - \widehat{u}_{jk}^{(\ell)}}{h}\right) \right\}, \quad (4.7)$$

where $(\widehat{u}_{ik}^{(1)}, \widehat{u}_{jk}^{(1)}) = (\widehat{u}_{ik}, \widehat{u}_{jk})$, $(\widehat{u}_{ik}^{(2)}, \widehat{u}_{jk}^{(2)}) = (-\widehat{u}_{ik}, \widehat{u}_{jk})$, $(\widehat{u}_{ik}^{(3)}, \widehat{u}_{jk}^{(3)}) = (\widehat{u}_{ik}, -\widehat{u}_{jk})$, $(\widehat{u}_{ik}^{(4)}, \widehat{u}_{jk}^{(4)}) = (-\widehat{u}_{ik}, -\widehat{u}_{jk})$, $(\widehat{u}_{ik}^{(5)}, \widehat{u}_{jk}^{(5)}) = (\widehat{u}_{ik}, 2 - \widehat{u}_{jk})$, $(\widehat{u}_{ik}^{(6)}, \widehat{u}_{jk}^{(6)}) = (-\widehat{u}_{ik}, 2 - \widehat{u}_{jk})$, $(\widehat{u}_{ik}^{(7)}, \widehat{u}_{jk}^{(7)}) = (2 - \widehat{u}_{ik}, \widehat{u}_{jk})$, $(\widehat{u}_{ik}^{(8)}, \widehat{u}_{jk}^{(8)}) = (2 - \widehat{u}_{ik}, -\widehat{u}_{jk})$, $(\widehat{u}_{ik}^{(9)}, \widehat{u}_{jk}^{(9)}) = (2 - \widehat{u}_{ik}, 2 - \widehat{u}_{jk})$. The kernel function $K(\cdot)$ is used to construct the muliplicative kernel $K(u_i, u_j) = K(u_i)K(u_j)$ and $h$ is the corresponding bandwidth. To convert the data $\{\boldsymbol{x}_i\}_{i=1}^n$ into the rank statistics $\{\widehat{\boldsymbol{u}}_i\}_{i=1}^n$, we squeeze the unbounded random vector $\boldsymbol{X}$ into the unit square $[0, 1]^2$ as shown in Figure 1. As there are fewer data points on the boundary than in the middle, the mirror reflection in (4.7) removes the boundary bias. Since all rank statistics locate on the grids, the kernel copula density estimator $\widetilde{c}_h$ is more robust than the classical kernel density estimator.

To keep our estimated copula density away from zero and infinity, we truncate the $\widetilde{c}_h(u_i, u_j)$ in (4.7) as follows

$$\widehat{c}_h(u_i, u_j) = \begin{cases} \kappa_1 & \text{if } \widetilde{c}_h(u_i, u_j) < \kappa_1; \\ \widetilde{c}_h(u_i, u_j) & \text{if } \kappa_1 \leq \widetilde{c}_h(u_i, u_j) \leq \kappa_2; \\ \kappa_2 & \text{if } \widetilde{c}_h(u_i, u_j) > \kappa_2. \end{cases} \quad (4.8)$$

The constants $\kappa_1, \kappa_2$ will be determined later according to the density function. Hence we have the mutual information estimator:

$$I_{ij}(\widehat{c}_h) = -H_{ij}(\widehat{c}_h) = \iint_{[0,1]^2} \widehat{c}_h(u_i, u_j) \log \widehat{c}_h(u_i, u_j) du_i du_j, \quad (4.9)$$

for each pair $i, j = 1, \ldots, d$ and obtain a $d \times d$ mutual information matrix $\widehat{M} = [I_{ij}(\widehat{c}_h)]$.

### 4.2.2 Mutual Information Concentration Inequality

For any $i, j = 1, \ldots, d$, the marginal bivariate density function of $(X_i, X_j)$ is $p(x_i, x_j)$. Let $c(u_i, u_j)$ be the copula density of $p(x_i, x_j)$, and it satisfies the following assumptions.

**Assumption 4.3** (Density assumption). We assume the density $c(u_i, u_j)$ belongs to a compact supported 2nd-order Hölder class $\Sigma_\kappa(2, L)$ and is bounded away from zero and infinity. The copula density satisfies

1. (Boundedness) There exist constants $\kappa_1, \kappa_2$ such that
$$0 < \kappa_1 \leq \min_{(u_i, u_j) \in [0,1]^2} c(u_i, u_j) \leq \max_{(u_i, u_j) \in [0,1]^2} c(u_i, u_j) \leq \kappa_2 < \infty;$$

2. (2nd-order Hölder class) For any $(u_i, u_j)^T \in [0, 1]^2$, there exists a constant $L$ such that, for any $s, t > 0$, if $\partial c/\partial u$ and $\partial c/\partial v$ are the first and second partial derivative of the copula density $c(u, v)$, then
$$\left| c(u_i + s, u_j + t) - c(u_i, u_j) - \frac{\partial c(u_i, u_j)}{\partial u} s - \frac{\partial c(u_i, u_j)}{\partial v} t \right| \leq L(s^2 + t^2);$$

3. (Boundary Assumption) The partial derivative of copula density $c(u_i, u_j)$ decreases to zero on the boundary, i.e., for any sequence $\{u_i^{(n)}, u_j^{(n)}\}_{n=1}^\infty \in [0, 1]^2$ converges to some point on the boundary, we have
$$\lim_{n \to \infty} \frac{\partial c(u_i^{(n)}, u_j^{(n)})}{\partial u} = \lim_{n \to \infty} \frac{\partial c(u_i^{(n)}, u_j^{(n)})}{\partial v} = 0.$$

We also need a kernel function $K(\cdot)$ satisfying the following assumptions:

**Assumption 4.4** (Kernel assumption). The kernel function satisfies

1. (Symmetry) The kernel $K(\cdot)$ is a symmetric and continuous density function with bounded support $[-1, 1]$.

2. (Lipschitz) For any $u_1, u_2 \in [-1, 1]$, the kernel $K(u)$ has the Liphschitz continuity
$$|K(u_1) - K(u_2)| \leq L_K |u_1 - u_2|.$$

For the mutual information estimator we proposed in (4.9), we have the following exponential concentration inequality. The proof is deferred to Appendix F.

**Theorem 4.5.** For a bivariate random vector $(X_i, X_j)$ with the copula density satisfying Assumption 4.3. The kernel copula density estimator $\widehat{c}_h$ in (4.8) using kernel function $K(\cdot)$ satisfying Assumption 4.4. Let $I_{ij}^* = I(X_i; X_j)$ be the population mutual information between $X_i, X_j$ and $I_{ij}(\widehat{c}_h)$ be the estimator proposed in (4.9). If we choose the bandwidth $h \asymp (\log n/n)^{1/6}$, then for sufficiently large $n$,
$$\sup_{c \in \Sigma_\kappa(2, L)} \mathbb{P}\left( |I_{ij}(\widehat{c}_h) - I_{ij}^*| > \epsilon \right) \leq 2 \exp\left( -\frac{(n^2 \log n)^{1/3} \epsilon^2}{128 \kappa^2 L_K^2} \right),$$
where $\kappa = \max\{|\log \kappa_1|, |\log \kappa_2|\} + 1$.

According to Chebyshev's inequality, we can easily get the upper bound of the absolute error from Theorem 4.5

**Corollary 4.6.** Under Assumptions 4.3 and 4.4, we have

$$\mathbb{E}\left(|I_{ij}(\widehat{c}_h) - I^*_{ij}|\right) = O\left(n^{-1/3} \log^{1/6} n\right).$$

**Remark 4.7.** In Liu et al. (2011, 2012c), the mutual information is estimated based on the classical kernel density estimators according to (2.5). Both the bivariate density functions and marginal univariate functions should be estimated. The corresponding estimator is

$$I(\widehat{p}_h) := \iint \widehat{p}_h(x_1, x_2) \log\left(\frac{\widehat{p}_h(x_1, x_2)}{\widehat{p}_h(x_1)\widehat{p}_h(x_2)}\right) dx_1 \, dx_2. \tag{4.10}$$

The bivariate density function estimator $\widehat{p}_h(x_1, x_2)$ is obtained by mirror reflection kernel density estimator similar to (4.7)

$$\widetilde{p}_h(x_i, x_j) := \frac{1}{nh^2} \sum_{k=1}^{n} \sum_{\ell=1}^{9} \left\{ K\left(\frac{x_i - x_{ik}^{(\ell)}}{h}\right) K\left(\frac{x_j - x_{jk}^{(\ell)}}{h}\right) \right\}, \tag{4.11}$$

where $(x_{ik}^{(1)}, x_{jk}^{(1)}) = (x_{ik}, x_{jk})$, $(x_{ik}^{(2)}, x_{jk}^{(2)}) = (-x_{ik}, x_{jk})$, $(x_{ik}^{(3)}, x_{jk}^{(3)}) = (x_{ik}, -x_{jk})$, $(x_{ik}^{(4)}, x_{jk}^{(4)}) = (-x_{ik}, -x_{jk})$, $(x_{ik}^{(5)}, x_{jk}^{(5)}) = (x_{ik}, 2 - x_{jk})$, $(x_{ik}^{(6)}, x_{jk}^{(6)}) = (-x_{ik}, 2 - x_{jk})$, $(x_{ik}^{(7)}, x_{jk}^{(7)}) = (2 - x_{ik}, x_{jk})$, $(x_{ik}^{(8)}, x_{jk}^{(8)}) = (2 - x_{ik}, -x_{jk})$, $(x_{ik}^{(9)}, x_{jk}^{(9)}) = (2 - x_{ik}, 2 - x_{jk})$, and $\widehat{p}_h(x_i, x_j)$ is derived by truncation as in (4.8). The univariate density estimator is derived following the one dimension version of (4.11). Here we choose the bandwidth $h \asymp n^{-1/4}$ for bivariate and univariate densities in Liu et al. (2012c). Then the entropy estimator for each node $\ell$ is obtained by

$$\widehat{H}_p(x_\ell) = -\int \widehat{p}_h(x_\ell) \log \widehat{p}_h(x_\ell) dx_\ell. \tag{4.12}$$

If we compare $I(\widehat{p}_h)$ in (4.10) with the copula-based estimator $I(\widehat{c}_h)$ in (4.9), for $I(\widehat{p}_h)$ we need to estimate both the bivariate density $\widehat{p}_h(x_i, x_j)$ and univariate densities $\widehat{p}_h(x_i), \widehat{p}_h(x_j)$. For $I(\widehat{c}_h)$, we only need to estimate bivariate copula density $\widehat{c}_h(u_i, u_j)$. Our estimator is more efficient in computation.

**Remark 4.8.** Using the bandwidth $h \asymp n^{-1/4}$ in (4.11), a corresponding concentration inequality has been proved by Liu et al. (2012c) for $I(\widehat{p}_h)$. If the density $p(x_i, x_j)$ is supported on a unit square $[0,1]^2$ and $p \in \Sigma_\kappa(2, L)$, we have

$$\sup_{p \in \Sigma_\kappa(2,L)} \mathbb{P}\Big(|I(\widehat{p}_h) - I(p)| > \epsilon\Big) \leq 6 \exp\left(-\frac{n\epsilon^2}{324\kappa^2}\right). \tag{4.13}$$

Comparing (3.3) with (4.13), our estimator has a slower rate of convergence. However, our estimator is robust to outliers and does not require the density function has a compact support on $[0,1]^2$ as required by Liu et al. (2012c).

### 4.2.3 Graph Estimation Consistency

In this section, we use the negative mutual information as the Fermat metric. and establish the graph estimation consistency of Algorithm 2. More specifically, we use the copula mutual information estimator (4.9) to estimate the Fermat metric and plug it into Algorithm 2. We call this method as Kernel Information-Theoretical Estimator (KITE). The following theorem shows the consistency of KITE.

**Theorem 4.9.** Suppose the graph metric $d_{ij} = -I(X_i; X_j)$ is a Fermat metric for the nonparametric graphical model. Under Assumptions 4.3, 4.4, and the assumption on the gap of the crucial set:

$$\min_{(e,e') \in \mathcal{C}} |d_e - d_{e'}| \geq L_n, \text{ where } L_n = \Omega\left(\sqrt{\frac{\log d + \log n}{(n^2 \log n)^{1/3}}}\right), \quad (4.14)$$

if we choose bandwidth $h \asymp (\log n/n)^{1/6}$ for the mutual information estimator $\widehat{I}_{ij}(\widehat{c}_h)$ in (4.9), then the estimated $\widehat{G}$ derived from Algorithm 2 with $\widehat{d}_{ij} = -\widehat{I}_{ij}(\widehat{c}_h)$ as input has $\lim_{n \to \infty} \mathbb{P}\left(\widehat{G} \neq G\right) = 0$.

*Proof.* Combining Theorem 4.5 and (4.14) which satisfies Assumption 3.5 for $r_n = (n^2 \log n)^{1/3}$, the graph estimation consistency follows from Theorem 3.6. □

### 4.2.4 Skeleton Forest Density Estimation

Let $p(\boldsymbol{x})$ be the true density that generates the data. In this section, our goal is to estimate the skeleton forest density $q^*$ defined as

$$q^* = \arg\min_{q \in \mathcal{P}_d} D(p||q),$$

where $D(p||q)$ is the Kullback-Leibler divergence between $p$ and $q$, and $\mathcal{P}_d$ is the family of distributions satisfying the tree factorization (2.4). Under the assumptions in Section 4.2.3, we can use Algorithm 2 to estimate a graph $G$ with loops. However, it is hard to estimate the density $p$ due to the curse of dimensionality.

In this section, we present an algorithm that estimates a skeleton forest density $q^*$ in the form of (2.4). Such a skeleton forest density provides the best tree approximation to the true density $p(\boldsymbol{x})$. To estimate the skeleton forest $q^*$, Liu et al. (2012c) propose to first estimate the forest graph structure of $q^*$ by MST and plug in the bivariate and univariate kernel density estimators into (2.4) to estimate $q^*$. However, their method relies on the assumption that $q^*$ has a bounded support. In this, we propose an estimator of the skeleton forest densities with unbounded supports.

Let $F^*$ be the graph of the forest density $q^*$ and we call it the skeleton forest. We estimate $F^*$ by $\widehat{F} = (V, \widehat{E})$, which is the output of Algorithm 1 with $\widehat{d}_{ij} = -\widehat{I}_{ij}(\widehat{c}_h)$ for $i, j = 1, \ldots, d$ as the input. Denote $\widehat{S}$ as the set of vertices isolated from others in the forest $\widehat{F}$. Let $\widehat{p}_{h_1}(x_\ell)$ be the standard kernel density estimator in (4.4) for all $\ell \in \widehat{S}$. We choose $h_1 \asymp n^{-1/5}$. The bivariate density estimator $\widehat{p}_{h_2}(x_j, x_k)$ and the univariate density estimator $\widehat{p}_{h_2}(x_j)$ for these nonisolated nodes $j, k$ are estimated by choosing $h_2 \asymp n^{-1/6}$ for the kernel density estimator in (4.4). Then, we obtain the skeleton forest density estimator

$$\widehat{p}_{\widehat{F}}(x) = \prod_{(j,k) \in \widehat{E}} \frac{\widehat{p}_{h_2}(x_j, x_k)}{\widehat{p}_{h_2}(x_j)\widehat{p}_{h_2}(x_k)} \cdot \prod_{u \in \widehat{S}} \widehat{p}_{h_1}(x_u) \cdot \prod_{\ell \in V \setminus \widehat{S}} \widehat{p}_{h_2}(x_\ell). \quad (4.15)$$

The following theorem presents the estimation rate of the skeleton forest density estimator (4.15), whose proof is provided in Appendix G.

**Theorem 4.10** (**Density Estimation Consistency**). Under Assumptions 4.3 and 4.4, if $\widehat{p}_{\widehat{F}}$ in (4.15) is obtained by choosing $h_1 \asymp n^{-1/5}$, $h_2 \asymp n^{-1/6}$ and $\log d / (n^2 \log n)^{1/3} = o(1)$, there exists a constant $C$ such that

$$\mathbb{E} \left\| \widehat{p}_{\widehat{F}} - q^* \right\|_{L_1} \leq C \sqrt{\frac{s}{n^{2/3}} + \frac{d-s}{n^{4/5}}},$$

where for any function $f(x)$, its $L_1$ norm is defined as $\|f\|_{L_1} = \int |f(x)| dx$.

**Remark 4.11.** The $L_1$ rate of convergence in Theorem 4.10 for density functions with arbitrary supports matches the one for bounded support density functions in Liu et al. (2012c) with a suboptimal scaling $\log d / (n^2 \log n)^{1/3} = o(1)$.

## 5 Experimental Results

In this section, we compare the numerical performance of KITE with other methods. We first introduce a pruning method to regularize the size of estimated graphs. Next, we compare KITE with three other methods: FDE, GLASSO and refit GLASSO. FDE (Liu et al., 2011) applies the Chow-Liu algorithm on the kernel density estimator in (4.9). GLASSO (Tibshirani, 1996) estimates the inverse covariance matrix under the Gaussian graphical model through $\ell_1$ regularization, and the refit GLASSO is a two-step method based on GLASSO: (1) obtaining an inverse covariance matrix by GLASSO and (2) refitting the model without $\ell_1$ regularization on the estimated support. At last, we apply these methods to a genomic dataset.

### 5.1 Graph Pruning

We introduce a graph pruning procedure to avoid overfitting. We divide the dataset $\mathcal{D} = (\boldsymbol{x}_1, \ldots, \boldsymbol{x}_n)$ into two parts: the training data $\mathcal{D}_1$ with size $n_1$ and the held-out data $\mathcal{D}_2$ with size $n_2$. We first use $\mathcal{D}_1$ to obtain mutual information estimators $\widehat{I}_{n_1}(X_i; X_j)$ for all $i, j \in V$ by (4.9) and the graph estimator $\widehat{G}_{n_1}$. On the held-out data $\mathcal{D}_2$, we also estimate the mutual information $\widehat{I}_{n_2}(X_i; X_j)$ for any $i, j \in V$ and estimate the marginal entropy by $\widehat{H}_{n_2}(X_k)$ in (4.12). More specifically, we prune the graph estimator $\widehat{G}_{n_1}$ by estimating the skeleton forest density. Let $\widehat{F}_{n_1}^{(k)}$ be the forest generated in the $k$-th iteration of Algorithm 1, and we choose the estimated $\widetilde{k}$ to be

$$\widetilde{k} = \underset{k \in \{0,1,\ldots,d\}}{\arg\max} \sum_{(i,j) \in E(\widehat{F}_{n_1}^{(k)})} \widehat{I}_{n_2}(X_i, X_j) - \sum_{\ell \in \bar{V}(\widehat{F}_{n_1}^{(k)})} \widehat{H}_{n_2}(X_\ell), \tag{5.1}$$

where $E(\widehat{F}_{n_1}^{(k)})$ is the edge set of $\widehat{F}_{n_1}^{(k)}$ and $\bar{V}(\widehat{F}_{n_1}^{(k)})$ is the set of non-isolated vertices of $\widehat{F}_{n_1}^{(k)}$. Note that the maximizer of (5.1) may not be unique since the value of the objective function may not change for some iterations. In general, $\widetilde{k}$ in (5.1) is a set of maximizers and we choose $\widehat{k} = \max \widetilde{k}$ as the iteration step to prune. We choose the graph $\widehat{G}_{n_1}^{(\widehat{k})}$ in the $\widehat{k}$-th iteration of Algorithm 2 as the pruned graph estimator.

## 5.2 Synthetic Data

We consider four types of graph construction: hub, constellation, band and cluster to generate inverse covariance matrices $\mathbf{\Sigma}^{-1}$. An illustration of these four graph patterns is shown in Figure 2 and Figure 3. All patterns are composed of several small connected components with special structures. The hub graph can be decomposed into stars. In each star, one node is chosen as the hub and all the others are connected to the hub. The constellation graph consists of subgraphs with at most one cycle. The subgraphs of band graph are bands such that each pair of nodes with coordinates $i, j$ in one band is connected if $|i - j| \leq 2$. The cluster graph is partitioned into several disconnected Erdös-Renyi random subgraphs with probability parameter 0.2. The support of $\mathbf{\Sigma}^{-1}$ is determined by these graph structures. The off-diagonal entries on the support of $\mathbf{\Sigma}^{-1}$ are uniformly generated from $[-30, 10]$. The diagonal entries are chosen such that the smallest eigenvalue of $\mathbf{\Sigma}$ is 1.

Based on the covariance matrix $\mathbf{\Sigma}$, we generate $n_1 = n_2 = 100$ samples of the $d$-dimensional Gaussian data $\boldsymbol{x}_1, \ldots, \boldsymbol{x}_n \sim N(\boldsymbol{\mu}, \mathbf{\Sigma})$ with $d = 100$, mean vector $\boldsymbol{\mu} = (0.5, \ldots, 0.5)$. To test the robustness of estimators, we distort the data by applying the Box-Cox transformation $y_{ik} = (\text{sign}(x_{ik})|x_{ik}|^\nu - 1)/\nu$ (Box and Cox, 1964) for $i = 1, \ldots, d$ and $k = 1, \ldots, n$. In the simulation, we choose $\nu = 2.5$ and obtain distorted Gaussian data $\boldsymbol{y}_1, \ldots, \boldsymbol{y}_n$. The third nonparanormal dataset is generated by letting $z_{ik} = F_i(x_{ik})$ for $i = 1, \ldots, d$ and $k = 1, \ldots, n$, where $F_i(\cdot)$ is the marginal distribution function of $\boldsymbol{X}_i$ for $i = 1, \ldots, d$. This transformation keeps the graph structure of $(\boldsymbol{X}_1, \ldots, \boldsymbol{X}_d)$ while the transformed data no longer follow the Gaussian distribution (Liu et al., 2009). When applying FDE, we use the mutual information estimator $I(\widehat{p}_h)$ in (4.9) and apply MST to obtain the estimated forest $\widehat{F}$.

The log-likelihood function on the held-out dataset $\mathcal{D}_2$ is applied to compare the performance of density estimators by KITE, FDE, GLASSO and refit GLASSO. For KITE with the estimated skeleton forest $\widehat{F} = (V, \widehat{E})$ and the isolated nodes $\widehat{S}$, the log-likelihood function is

$$\ell_{\text{kite}} = \frac{1}{n_2} \sum_{s \in \mathcal{D}_2} \log \left( \prod_{(j,k) \in \widehat{E}} \frac{\widetilde{p}_{h_2}(X_j^{(s)}, X_k^{(s)})}{\widetilde{p}_{h_2}(X_j^{(s)}) \widetilde{p}_{h_2}(X_k^{(s)})} \cdot \prod_{u \in \widehat{S}} \widetilde{p}_{h_1}(X_u^{(s)}) \cdot \prod_{\ell \in V \setminus \widehat{S}} \widetilde{p}_{h_2}(X_\ell^{(s)}) \right),$$

and the held-out log-likelihood function of FDE is

$$\ell_{\text{fde}} = \frac{1}{n_2} \sum_{s \in \mathcal{D}_2} \log \left( \prod_{(j,k) \in \widehat{E}} \frac{\widetilde{p}_{h_2}(X_j^{(s)}, X_k^{(s)})}{\widetilde{p}_{h_2}(X_j^{(s)}) \widetilde{p}_{h_2}(X_k^{(s)})} \cdot \prod_{u \in V_{\widehat{F}}} \widetilde{p}_{h_2}(X_u^{(s)}) \right).$$

For GLASSO and refit GLASSO, let $\widehat{\boldsymbol{\mu}}_{n_1}, \widehat{\boldsymbol{\Omega}}_{n_1}$ be the estimated mean and inverse covariance matrix using $\mathcal{D}_1$, the held-out log-likelihood function is

$$\ell_{\text{gauss}} = \frac{1}{n_2} \sum_{s \in \mathcal{D}_2} \left\{ \frac{1}{2}(X^{(s)} - \widehat{\boldsymbol{\mu}}_{n_1})^T \widehat{\boldsymbol{\Omega}}_{n_1} (X^{(s)} - \widehat{\boldsymbol{\mu}}_{n_1}) + \frac{1}{2} \log \left( \frac{|\widehat{\boldsymbol{\Omega}}_{n_1}|}{(2\pi)^d} \right) \right\}.$$

We also tune the regularization parameter of GLASSO using the log-likelihood function above on the held-out data.

The graph estimation results for various graph patterns are shown in Figure 2 and Figure 3. We observe that GLASSO tends to over-select edges between two disconnected subgraphs. FDE tends to achieve a graph estimator sparser than the true graph, because it enforces the forest structure. On the one hand, this helps FDE to discover the connected structures, but on the other hand, the forest structure makes the FDE graph estimator have many false negative edges. KITE balances the performance of detecting the connected structures and exploring the structure for each subgraph. For the distorted datasets, simulations demonstrate that KITE is more robust than GLASSO and FDE.

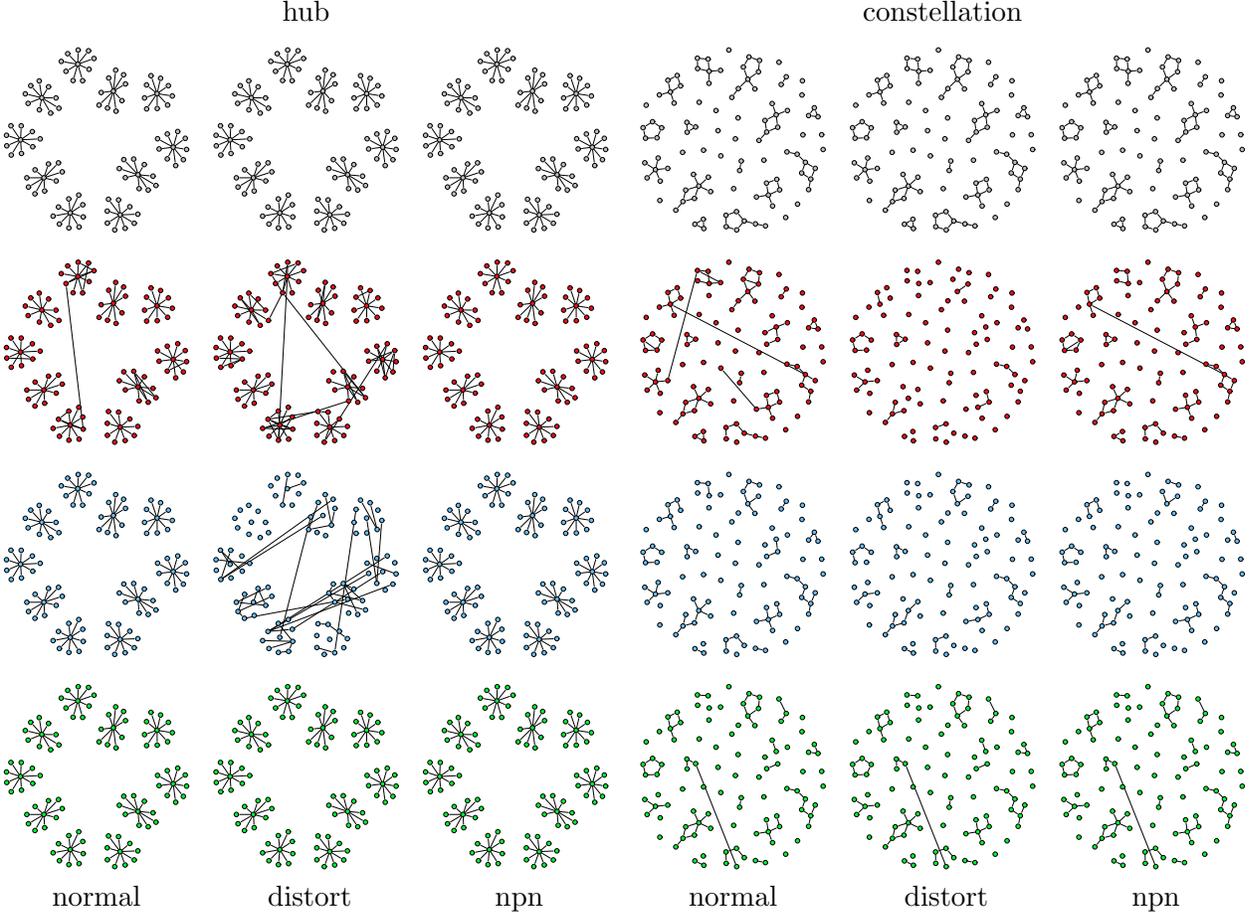

Figure 2: True and estimated graph structure for hub and forest graph models. For each graph pattern, the true graph (first row) is compared with the estimated graphs by GLASSO (second row), FDE (third row) and KITE (forth row) on Gaussian data (first column), distorted Gaussian data (second column) and nonparanormal data (third column).

From the values of held-out log-likelihood functions shown in Figure 4, the refit GLASSO is better than the other three methods on the Gaussian datasets because it uses the correct model. For the nonparanormal distribution data, although neither the Gaussian graphical model nor the forest density is correct, KITE and FDE have larger held-out log-likelihood function values than GLASSO and the refit GLASSO. This observation implies that the nonparametric forest density family have better approximation than the Gaussian graphical model. We see that KITE has similar performance on Gaussian dataset as FDE but outperforms FDE on the nonparanormal datasets.

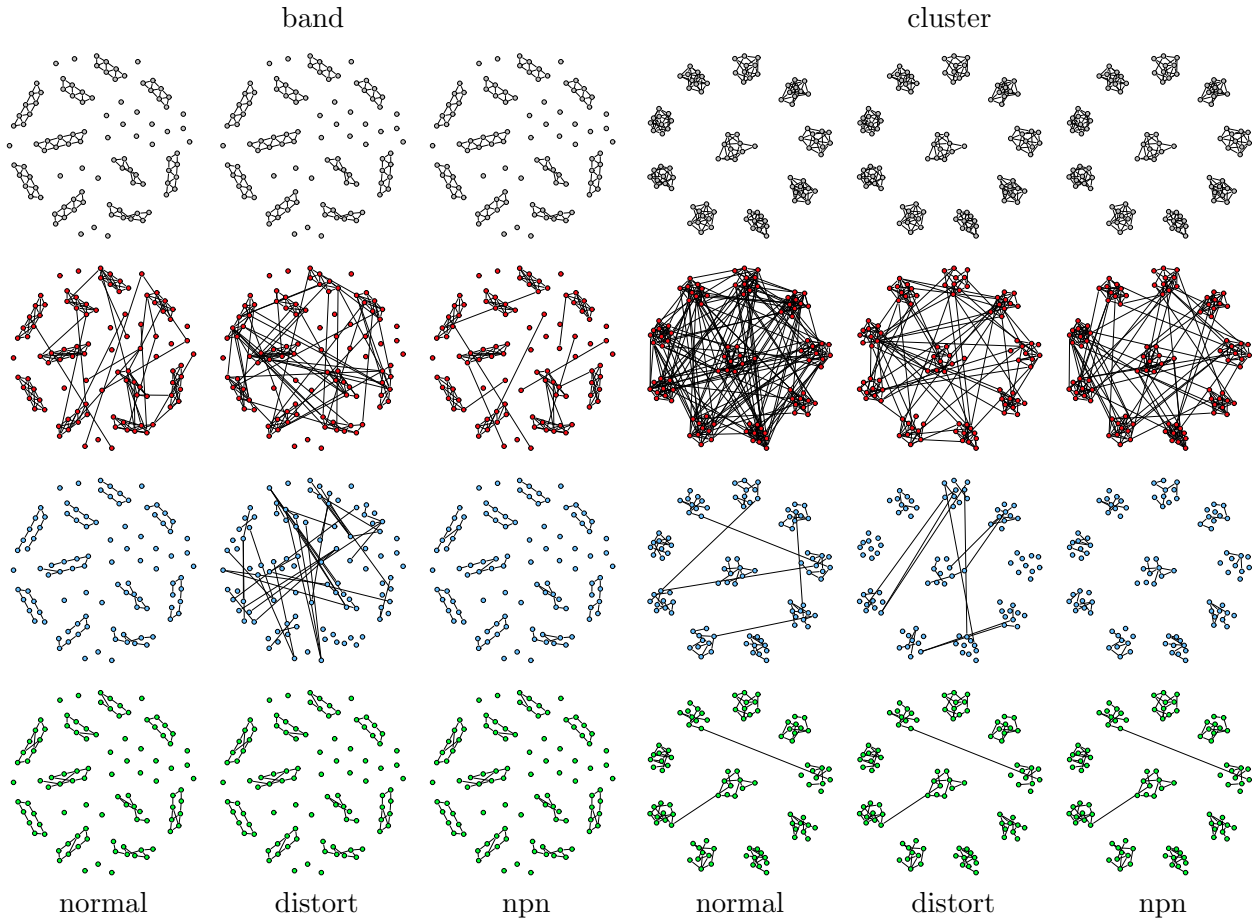

Figure 3: The true and estimated graph structures for band and cluster graph models. For each graph pattern, true graph (first row) is compared with estimated graphs by GLASSO (second row), FDE (third row) and KITE (forth row) on Gaussian data (first column), distorted Gaussian data (second column) and nonparanormal data (third column).

## 5.3 Gene Network

We apply the graph estimation methods on the genomic dataset in Wille et al. (2004). The dataset consists of gene expression arrays for Arabidopsis thaliana from $n = 118$ samples. We focus on $d = 39$ genes: 16 in the cytoplasm from the mevalonate (MVA) pathway, 18 in the chloroplast from the plastidial (MEP) pathway, 5 in the mitochondria. We first take a log-transformation and standardize the original gene expression level data; KITE, FDE and GLASSO methods are then applied to the preprocessed data to estimate the gene network.

The estimated gene networks are demonstrated in Figure 5. Even though these genes are located in different pathways, all the estimated graphs from these three methods are connected. This shows that there are cross-pathway links which is consistent to the conclusion in Wille et al. (2004). We can see that the gene network derived from KITE both reveals the cycle structures and has a more concise edge selection than GLASSO.

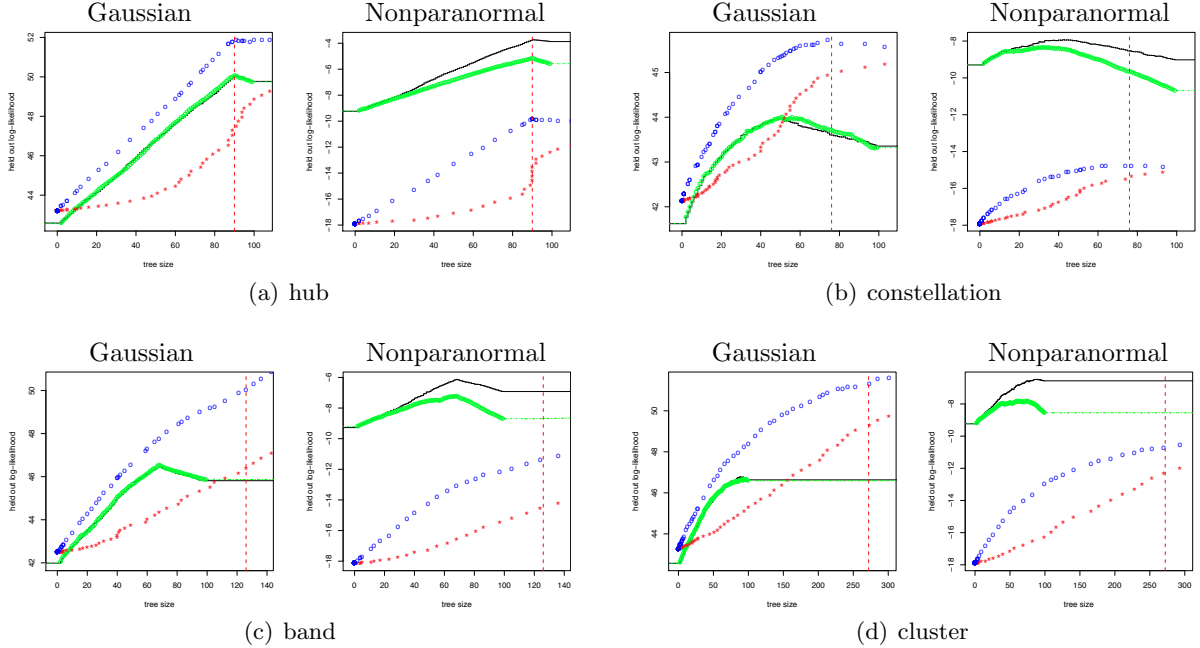

Figure 4: The held-out log-likelihood functions for GLASSO (blue circle), refit GLASSO (red star), FDE (green circle dash line) and KITE (black step function) on the Gaussian (left) and nonparanormal (right) datasets under the hub, constellation, band and cluster graph patterns. The red dash vertical line marks the true graph size.

# 6 Conclusion

We propose the graphical Fermat's principle as a new regularization on graphical models. The consistency of density estimation and graph estimation are proved under the exponential concentration inequality of the Fermat metric estimator. Concrete examples on the discrete and nonparametric graphical models are discussed. For nonparametric graph estimation, we also propose a robust copula-based mutual information estimator. Numerical experiments show that our estimator can robustly estimate the loopy graphs even on the contaminated data and outperform many existing graph estimators.

# A Proof of Theorem 2.10

We first prove that the local tree metric $\mathbf{d}_{\text{tree}}(V) = d(i, j; \text{tree}) = -\log|\det P_{\mathbf{X}_{i,j}|\text{tree}(i,j)}|$ is a Fermat metric. According to the Fact 2 in the supplemental materials of Anandkumar and Valluvan (2013), $\mathbf{d}_{\text{tree}}(V)$ forms a tree metric on the subtree of $G$. Now we verify (2.6) is true for $\mathbf{d}_{\text{tree}}(V)$. For any $i, j \in V$, since the geodesic distance $m_{ij} \leq \lfloor \frac{g}{2} \rfloor$, we have $G(B_l(i) \cup B_l(j))$ is a tree and $\text{Path}_{i,j}(m_{ij} + 1) \subset \text{tree}(i, j)$. Therefore, if two vertices $u, v$ with $\text{Path}_{u,v}(m_{uv}) \subsetneq \text{Path}_{i,j}(m_{ij})$, we have $\text{Path}_{u,v}(m_{uv}) \subset \text{tree}(i, j)$ as well. As the distances $\mathbf{d}_{\text{tree}}(B_l(i) \cup B_l(j))$ forms a tree metric, due

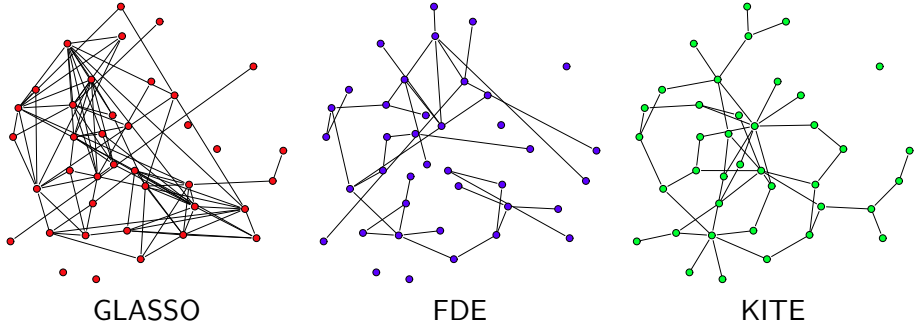

Figure 5: True and estimated gene network structure by GLASSO (top left), FDE (top middle) and KITE (top right).

to the additivity of the tree metric (2.8), we have

$$d(i,j;\text{tree}) = \sum_{(s,t)\in \text{Path}_{i,j}(m_{ij})} d(s,t;\text{tree}) < \sum_{(s,t)\in \text{Path}_{u,v}(m_{uv})} d(s,t;\text{tree}) = d(u,v;\text{tree}).$$

Similarly, combining the facts that $\text{Path}_{i,j}(m_{ij}+1) \subset G(B_l(i) \cup B_l(j))$ and $\mathbf{d}_{\text{tree}}(B_l(i) \cup B_l(j))$ forms a tree metric, for any $(u,v) \in \text{Path}_{i,j}(m_{ij}+1)$, it can be derived that $d(i,j;\text{tree}) = \sum_{(s,t)\in \text{Path}_{i,j}(m_{ij}+1)} d(s,t;\text{tree}) > d(u,v;\text{tree})$ which verifies (2.7).

Now we turns to prove the information distance $d_{ij} = -\log|\det(P_{\mathbf{X}_{i,j}})|$ is a Fermat metric. If the correlation decay in (2.11) is satisfied, from Proposition 2 in the supplemental materials of Anandkumar and Valluvan (2013), we have $|d_{ij} - d(i,j;\text{tree})| \leq |\mathcal{X}|e^{d'_{\max}(l)}\zeta(g/2-l-1)$. Under the condition that $\delta_{\max} \geq 2|\mathcal{X}|e^{d'_{\max}(l)}\zeta(g/2-l-1)$, $\{d_{ij}\}_{i,j\in V}$, the information distances preserve the relative order of the local tree metric $d(i,j;\text{tree})$. Because the graphical Fermat's principle is invariant under monotone transformations, we have $\{d_{ij}\}_{i,j\in V}$ is a Fermat metric.

### A.1 Correlation Decay under Ising Model

In this section, we discuss the connection between the correlation decay and Fermat principle under Ising model.

**Assumption A.1.** The following assumptions are considered by Anandkumar and Valluvan (2013) for the consistency of graph recovery of LocalCLGrouping for Ising models $P_G$.

1. The edge potentials satisfy $\theta_{\min} \leq |\theta_{i,j}| \leq \theta_{\max}$, for any $i,j \in V$.

2. We define the attractive counterpart $\bar{P}_G = \exp\left(\sum_{(i,j)\in E} |\theta_{ij}|x_i x_j + \sum_{i\in V} |\phi_i|x_i - A(|\boldsymbol{\theta}|)\right)$ and $\phi'_{\max} := \max_{i\in V} \text{atanh}(\bar{\mathbb{E}}(X_i))$, where the expectation $\bar{\mathbb{E}}$ is taken over $\bar{P}_G$. Let $P^{(1)}_{\mathbf{X}_{1,2}} = \exp(\theta_{\max}x_1 x_2 + \phi'_{\max}(x_1+x_2) - A(\boldsymbol{\theta_1}))$ and $P^{(2)}_{\mathbf{X}_{1,2}} = \exp(\theta_{\min}x_1 x_2 - A(\boldsymbol{\theta_2}))$ where $A(\boldsymbol{\theta_1}), A(\boldsymbol{\theta_2})$ are the normalizers. We define $d'_{\min} = -\log|\det(P^{(1)}_{\mathbf{X}_{1,2}})|$, $d'_{\max} = -\log|\det(P^{(2)}_{\mathbf{X}_{1,2}})|$ and $\eta := d'_{\max}/d'_{\min}$. Let $\Delta_{\max}$ be the maximum node degree and $\alpha := \Delta_{\max}\tanh\theta_{\max} < 1$. We assume

$$\frac{\alpha^{g/2}}{\theta_{\min}^{\eta(\eta+1)+2}} = o(1),$$

where $o(1)$ is with respect to the number of nodes $d$.

Therefore, we can derive the following result characterizing when the information distance $d_{ij}$ in Ising models will become a Fermat metric.

**Corollary A.2.** Let $p_G(\boldsymbol{x})$ be the density of Ising model in (2.3). Under Assumption A.1, if $\delta_{\max} \geq 2|\mathcal{X}|e^{d'_{\max}(l)}\zeta(g/2 - l - 1)$, the information distance $\{d_{ij}\}_{i,j \in V}$ is a Fermat metric.

*Proof.* In the proof of Theorem 1 in Anandkumar and Valluvan (2013), it is shown that under Assumption A.1, due to the property of Ising models, Assumption 2.9 is satisfied. Therefore, according to Theorem 2.10, $d_{ij} = -\log|\det(P_{\boldsymbol{X}_{i,j}})|$ is a Fermat metric. □

## B  Proof of Theorem 3.1

*Proof.* If the $k$-th edge $(i_k, j_k)$ is not added to $E_k^{\mathrm{g}}$, a new triangle must form in $E_k$ and we denote the third vertex of the triangle as $h$. Due to the algorithm, we have $(h, i_k) \in E_{k-1}$ and $(h, j_k) \in E_{k-1}$. We claim that there must exists a path connecting $h$ and $i_k$ in $\widehat{F}_{k-1}$. Suppose $(h, i_k)$ is considered in the $\ell$-th iteration of the algorithm and as $(h, i_k) \in E_{k-1}$, we have $\ell < k$. If $(h, i_k) \in E_k^{\mathrm{t}}$, $(h, i_k)$ is the path in the claim. If $(h, i_k) \notin E_k^{\mathrm{t}}$, a cycle must be formed in $E_\ell^{\mathrm{t}}$ in the $\ell$-th iteration when adding $(h, i_k)$. This implies that there is a path connecting $h, i_k$ in $\widehat{F}_\ell$. Similarly, it can also be derived that there is a path connecting $(h, j_k)$ in $\widehat{F}_{k-1}$ as well. Therefore, the edge $(i_k, j_k)$ will form a new cycle if it is added to $\widehat{F}_{k-1}$ and we have $(i_k, j_k) \notin E_k^{\mathrm{t}}$. We conclude that $E_k^{\mathrm{t}} \subseteq E_k$ for every $k = 1, \ldots, d$, and Algorithm 2 generates three sequences of graphs with the filtration structure in (3.1) □

## C  Proof of Lemma 3.4

*Proof.* We use the same notation as in Algorithm 2 and denote the fermat metric of any edge $e \in E$ as $d_e$. We will prove inductively that for every $k$-th iteration, the $e_k = (i_k, j_k)$ will be added to $E_k$ if and only if $e_k$ is the edge in true graph $G$. Let $e_1 = (i_1, j_1)$ be the edge with smallest graph metric, which is considered in the first iteration, $e_1$ must be a true edge, otherwise according to (2.6) there must exists $e' \in \mathrm{Path}(i_1, j_1)$ and $d_{e'} < d_{e_1}$ which violates the fact that $d_{e_1}$ is the smallest.

For the $k$-th iteration and the edge $e_k \in E$, the induction hypothesis is $E_{k-1}^{\mathrm{g}} \subset E$. Suppose $E_{k-1} \cup \{e_k\}$ forms a new triangle, and we denote the other two edges of the triangle are $(i_k, u)$ and $(u, j_k)$. As $G$ is triangle-free, there must be at least one of these two edges, say $(u, j_k) \notin E$ such that $\mathrm{Path}(i_k, u) \neq \varnothing$ (otherwise $d_{(i_k;u)} = d_{(u;j_k)} = \infty$ contradicting the fact that $d_{(u,j_k)} < d_{e_k}$). Let $\mathrm{path}(i_k, u)$ be a path connecting $u, i^k$, we construct a $\mathrm{path}(j_k, u) = \mathrm{path}(i_k, u) \cup \{(i_k, j_k)\} \subset \mathrm{Path}(u, j_k)$ while $d_{(u,j_k)} > d_{(i_k,j_k)}$ which contradicts (2.7). Therefore, if $e_k \in E$, it will not form a new triangle and hence be added into $E_{(k-1)}$.

If $e_k \notin E$, let the path connecting $i_k, j_k$ be $\mathrm{path}(i_k, j_k) \subset E$. Find a node $v \in V$ such that $(v, i_k) \in \mathrm{Path}(i_k, j_k)$, and it is easy to see that $\mathrm{Path}(v, j_k) \subsetneq \mathrm{Path}(i_k, j_k)$. According to (2.6), we have $d_{(v,i_k)} < d_{e_k}$ and $d_{(v,j_k)} < d_{e_k}$. This implies that the edges $(v, i_k), (v, j_k) \in E_{k-1}$ and $e_k$ will not be added to $E_k^{\mathrm{t}}$ as a new triangle is formed in $E_k = E^{k-1} \cup e_k$.

In conclusion, previous induction reasoning demonstrate that $\widehat{G}_{d^2} = G$. □

# D  Proof of Theorem 3.6

*Proof.* Since the output of Algorithm 2 only depends on the relative order of mutual information, we have

$$\mathbb{P}\left(\widehat{G} \neq G^*\right) \leq \mathbb{P}\left(\text{sign}(d_e - d_{e'}) \neq \text{sign}(\widehat{d}_e - \widehat{d}_{e'}) \text{ for some } e, e' \in E\right)$$
$$= \mathbb{P}\left((d_e - d_{e'}) \cdot (\widehat{d}_e - \widehat{d}_{e'}) < 0 \text{ for some } e, e' \in E\right)$$
$$\leq \frac{d^4}{2} \max_{(e,e') \in \mathcal{C}} \mathbb{P}\left((d_e - d_{e'}) \cdot (\widehat{d}_e - \widehat{d}_{e'}) < 0 \text{ for some } e, e' \in E\right).$$

Here the first inequality is because the relative orders of edges are same if the estimated graph is same as the truth. The last inequality is according to union bound. According to Assumption 3.5 we have $\min_{(e,e') \in \mathcal{C}} |d_e - d_{e'}| > 2L_n$, where $L_n = \Omega(\sqrt{(\log d + \log n)/r_n})$. Thus

$$\max_{(e,e') \in \mathcal{C}} \mathbb{P}\left((d_e - d_{e'}) \cdot (\widehat{d}_e - \widehat{d}_{e'}) \text{ for some } e, e' \in E\right)$$
$$\leq \max_{e \in E} \mathbb{P}\left(|d_e - \widehat{d}_e| > 2L_n\right) \leq C_1 \exp\left(-C_2 L_n^2 r_n\right),$$

and the last inequality follows from (3.3). Combining the above two inequalities, we can derive that

$$\mathbb{P}\left(\widehat{G} \neq G^*\right) \leq \frac{d^4}{2} \max_{(e,e') \in \mathcal{C}} \mathbb{P}\left((d_e - d_{e'}) \cdot (\widehat{d}_e - \widehat{d}_{e'}) < 0 \text{ for some } e, e' \in E\right)$$
$$\leq d^4 \max_{e \in E} \mathbb{P}\left(|d_e - \widehat{d}_e| > 2L_n\right)$$
$$\leq C_1 d^4 \exp\left(-C_2 L_n^2 r_n\right) = O\left(\exp\left(4 \log d - C_2 L_n^2 r_n\right)\right) = o(1). \tag{D.1}$$

Therefore, we complete the proof of the theorem. □

# E  Proof of Lemma 4.1

In order to show the concentration of the Fermat metric $\widehat{d}'_{ij} = -\exp(-\widehat{d}_{ij})$ for $\widehat{d}_{ij} = -\log|\widehat{P}_{ij}| + \frac{1}{2}\log|\widehat{P}_{ii}| + \frac{1}{2}\log|\widehat{P}_{jj}|$. We first bound the concentration determinant of the empirical stochastic matrix by the following lemma.

**Lemma E.1.** (Anandkumar and Valluvan, 2013) Let $\widehat{P} \in \mathbb{R}^{s \times s}$ be the empirical stochastic matrix based on the data with sample size $n$, and we have

$$\mathbb{P}\left(|\det(\widehat{P}) - \det P| > \epsilon\right) \leq 2^s \exp\left(-\frac{n\epsilon^2}{2s^2}\right).$$

Now we prove the concentration inequality (4.1). First, according to the definition of $\widehat{d}_{ij}$ and $d_{ij}$, we can derive that

$$\mathbb{P}(|\exp(-\widehat{d}_{ij}) - \exp(-d_{ij})| > \epsilon) = \mathbb{P}(|\det(\widehat{P}_{ii}^{-1/2} \widehat{P}_{ij} \widehat{P}_{jj}^{-1/2}) - \det(P_{ii}^{-1/2} P_{ij} P_{jj}^{-1/2})| > \epsilon) \leq T_1 + T_2 + T_3,$$

where the last decomposition is due to union bound such that

$$T_1 = \mathbb{P}\left(|\det(\widehat{P}_{ii}^{-1/2}) - \det(P_{ii}^{-1/2})| \cdot |\det(\widehat{P}_{ij})\det(\widehat{P}_{jj}^{-1/2})| > \epsilon/3\right);$$

$$T_2 = \mathbb{P}\left(|\det(\widehat{P}_{jj}^{-1/2}) - \det(P_{jj}^{-1/2})| \cdot |\det(P_{ii}^{-1/2})\det(\widehat{P}_{ij})| > \epsilon/3\right);$$

$$T_3 = \mathbb{P}\left(|\det(\widehat{P}_{ij}) - \det(P_{ij})| \cdot |\det(P_{ij}^{-1/2})\det(P_{jj}^{-1/2})| > \epsilon/3\right).$$

By triangle inequality and the fact that $\min_{i,j \in V} \|P_{ij}\|_{\min} > p_0$, we have

$$T_1 + T_2 + T_3 \leq \mathbb{P}\left(|\det(\widehat{P}_{ii}) - \det(P_{ii})| > \frac{p_0^{2s}\epsilon}{4}\right) + \mathbb{P}\left(|\det(\widehat{P}_{jj}) - \det(P_{jj})| > \frac{p_0^s}{2}\right)$$

$$+ \mathbb{P}\left(|\det(\widehat{P}_{ii}) - \det(P_{ii})| > \frac{p_0^{2s}\epsilon}{2}\right) + \mathbb{P}\left(|\det(\widehat{P}_{jj}) - \det(P_{jj})| > \frac{p_0^s}{2}\right)$$

$$+ \mathbb{P}\left(|\det(\widehat{P}_{ij}) - \det(P_{ij})| > \frac{p_0^s\epsilon}{2}\right).$$

Now we transfer the concentration of metric estimators to the concentration of the determinant of probability matrix. Therefore, according to Lemma E.1, we have

$$\mathbb{P}(|\exp(-\widehat{d}_{ij}) - \exp(-d_{ij})| > \epsilon) \leq T_1 + T_2 + T_3$$

$$\leq 3 \cdot 2^s \exp\left(-\frac{p_0^{4s}n\epsilon^2}{32s^2}\right) + 2^{s+1}\exp\left(-\frac{p_0^{2s}n}{8s^2}\right) \leq 2^{s+2}\exp\left(-\frac{p_0^{4s}n\epsilon^2}{32s^2}\right),$$

where the last inequality is derived from $\epsilon > \sqrt{32p_0^{-s}}$.

## F  Proof of Theorem 4.5

In this section, we will analysis the statistical properties of kernel copula density estimator and give a proof of Theorem 4.5. The error $|I(\widehat{c}_h) - I(c)|$ could be decomposed as two parts

$$|I(\widehat{c}_h) - I(c)| \leq |I(\widehat{c}_h) - \mathbb{E}I(\widehat{c}_h)| + |\mathbb{E}I(\widehat{c}_h) - I(c)|.$$

The first part on the right side of the inequality is bias term and the second part is variance term. We are going to analyze these two parts separately and show that

$$\textbf{Bias Term}: \quad \sup_{c \in \Sigma_\kappa(2,L)} |\mathbb{E}I(\widehat{c}_h) - I(c)| \leq C_1 \left(\frac{\log n}{n}\right)^{1/3};$$

$$\textbf{Variance Term}: \quad \sup_{c \in \Sigma_\kappa(2,L)} \mathbb{P}\Big(|I(\widehat{c}_h) - \mathbb{E}I(\widehat{c}_h)| > \epsilon\Big) \leq 2\exp\left(-\frac{(n^2\log n)^{1/3}\epsilon^2}{128\kappa^2 L_K^2}\right),$$

where $C_1$ is a constant only depend on the properties of copula density and kernel functions and the bandwidth is set to be $h \asymp (\log n/n)^{1/6}$.

## F.1 Bias Term Analysis

For the purpose of estimating the bias of mutual information, we first estimate the integrated squared bias of kernel copula density estimator in the following lemma.

**Lemma F.1.** Under Assumption 4.3 and 4.4, we have for some constant $c$,

$$\sup_{c \in \Sigma_\kappa(2,L)} \int_{[0,1]^2} (\mathbb{E}\widetilde{c}_h(u_i, u_j) - c(u_i, u_j))^2 \, du_i \, du_j \leq c \left( h^4 + \frac{\log n}{nh^2} + \frac{1}{n^8} \right).$$

*Proof.* As we apply mirror reflection kernel estimator in 4.7, the estimator's behavior is different on the boundary area and central area of the unit square. We will analyze the properties in various areas.

$$\int_{[0,1]^2} (\mathbb{E}\widetilde{c}_h(u_i, u_j) - c(u_i, u_j))^2 \, du_i \, du_j$$
$$= \int_{\mathcal{A}} + \int_{\mathcal{B}} + \int_{\mathcal{C}} (\mathbb{E}\widetilde{c}_h(u_i, u_j) - c(u_i, u_j))^2 \, du_i \, du_j =: \mathcal{T}_{\mathcal{A}} + \mathcal{T}_{\mathcal{B}} + \mathcal{T}_{\mathcal{C}},$$

where $\mathcal{A}$ is central area, $\mathcal{B}$ is the four corners and $\mathcal{C}$ is four margins defined as follows:

$$\begin{aligned}
\mathcal{A} &= [h, 1-h]^2 \\
\mathcal{B} &= [0,h]^2 \cup [1-h,1] \times [0,h] \cup [0,h] \times [1-h,1] \cup [1-h,1]^2 \\
\mathcal{C} &= [0,1]^2 \setminus (\mathcal{A} \cup \mathcal{B})
\end{aligned}$$

We substitute $\boldsymbol{u}_i = F^{(i)}(\boldsymbol{x}_i)$, $\boldsymbol{u}_j = F^{(i)}(\boldsymbol{x}_j)$ for $\widehat{\boldsymbol{u}}_i, \widehat{\boldsymbol{u}}_j$ in (4.7), we will get the standard kernel density estimator of copula density and we denote it as $\widetilde{c}_h^{\text{std}}(u_i, u_j)$.

### F.1.1 Analysis of Central Area $\mathcal{T}_{\mathcal{A}}$

For any $(u_i, u_j) \in \mathcal{A} = [h, 1-h]^2$, the mirror reflection kernel estimator is

$$\widetilde{c}_h(u_i, u_j) = \frac{1}{nh^2} \sum_{k=1}^n K\left(\frac{u_i - \widehat{u}_{ik}}{h}\right) K\left(\frac{u_j - \widehat{u}_{jk}}{h}\right).$$

According the formulation of $\widetilde{c}_h(u_i, u_j)$, we use the variance-bias decomposition and obtain that

$$\mathcal{T}_{\mathcal{A}} \leq 2 \int_{\mathcal{A}} \left( \mathbb{E}\widetilde{c}_h - \mathbb{E}\widetilde{c}_h^{\text{std}} \right)^2 du_i \, du_j + 2 \int_{\mathcal{A}} \left( \mathbb{E}\widetilde{c}_h^{\text{std}} - c \right)^2 du_i \, du_j =: 2(\mathcal{T}_{\mathcal{A}}^{(1)} + \mathcal{T}_{\mathcal{A}}^{(2)}),$$

where $\widetilde{c}_h^{\text{std}}$ is achieved by substituting $\boldsymbol{x}_i = F^{(i)}(\boldsymbol{x}_i)$, $\boldsymbol{x}_j = F^{(j)}(\boldsymbol{x}_j)$ for $\widehat{\boldsymbol{x}}_i, \widehat{\boldsymbol{x}}_j$ in (4.7). The second term is integrated square bias for standard kernel density, we have already known that $\sup_{p \in \Sigma(2,L)} \mathcal{T}_{\mathcal{A}}^{(2)} \leq ch^4$ (Silverman, 1986). To estimate $\mathcal{T}_{\mathcal{A}}^{(1)}$, denote $K_h(\cdot) = h^{-1}K(\cdot/h)$ and we plug in the constructions of copula estimators and have

$$
\iint_{\mathcal{A}} \left(\mathbb{E}\widetilde{c}_h - \mathbb{E}\widetilde{c}_h^{\text{std}}\right)^2 du_i\, du_j \leq \iint_{[0,1]^2} \left(\frac{1}{n}\sum_{k=1}^n \mathbb{E}\left|K_h\bigl(u_i - F_n^{(i)}(x_{ik})\bigr)K\bigl(u_j - F_n^{(j)}(x_{jk})\bigr)\right.\right.
$$
$$
\left.\left. - K_h\bigl(u_i - F^{(i)}(x_{ik})\bigr)K_h\bigl(u_j - F^{(j)}(x_{jk})\bigr)\right|\right)^2 du_i\, du_j
$$
$$
= \iint_{[0,1]^2} \left(\frac{1}{n}\sum_{k=1}^n \mathbb{E}\left|K\bigl(u_i - h^{-1}F_n^{(i)}(x_{ik})\bigr)K\bigl(u_j - h^{-1}F_n^{(j)}(x_{jk})\bigr)\right.\right.
$$
$$
\left.\left. - K\bigl(u_i - h^{-1}F^{(i)}(x_{ik})\bigr)K\bigl(u_j - h^{-1}F^{(j)}(x_{jk})\bigr)\right|\right)^2 du_i\, du_j,
$$

where the last equality is because of change of variable. We define $\Delta_k(x_i, x_j) = \max\{|F_n^{(i)}(x_{ik}) - F_n^{(i)}(x_{ik})|, |F_n^{(j)}(x_{jk}) - F_n^{(j)}(x_{jk})|\}$ and $C_k = 4\|K\|_\infty L_k$. Due to the Lipschitz continuity of $K(\cdot)$ in Assumption 4.4 and the above inequality, we have

$$
\mathcal{T}_{\mathcal{A}}^{(1)} \leq \left(\frac{1}{n}\sum_{k=1}^n 2\|K\|_\infty \mathbb{E}\left(L_K \frac{\epsilon}{h}\mathbb{1}\{\Delta_k(x_i, x_j) \leq \epsilon\} + \mathbb{1}\{\Delta_k(x_i, x_j) \geq \epsilon\}\right)\right)^2
$$
$$
\leq C_k^2 \left(\frac{\epsilon}{h} + \mathbb{P}\left\{\max_{1\leq k\leq n}\Delta_k(x_i, x_j) > \epsilon\right\}\right)^2 \leq 2C_k^2\left(\frac{\epsilon^2}{h^2} + e^{-4n\epsilon^2}\right), \quad \text{(F.1)}
$$

where the last inequality is derived by Dvoretzky-Kiefer-Wolfowitz inequality (Dvoretzky et al., 1956) that $\mathbb{P}\left(\sup_{x\in\mathbb{R}}|F_n(x) - F(x)| > \epsilon\right) \leq e^{-2n\epsilon^2}$ for any distribution function $F(\cdot)$.

Let $\epsilon = \sqrt{\frac{2\log n}{n}}$, we have $\mathcal{T}_{\mathcal{A}}^{(1)} \leq 2C_K(\log n/(nh^2) + n^{-8})$ and $T_{\mathcal{A}} \leq c[h^4 + \log n/(nh^2) + n^{-8}]$. Similar to the analysis of $\mathcal{T}_{\mathcal{A}}^{(1)}$, we can bound the mean square error using the same $\epsilon$ as

$$
\mathcal{T}_{\mathcal{A}}^{(2)} \leq \iint_{[0,1]^2} \mathbb{E}\left(\frac{1}{n}\sum_{k=1}^n \left|K\bigl(u - F_n^{(i)}(x_{ik})/h\bigr)K\bigl(v - F_n^{(j)}(x_{jk})/h\bigr)\right.\right.
$$
$$
\left.\left. - K\bigl(u - F_n^{(i)}(x_{ik})/h\bigr)K\bigl(v - F_n^{(j)}(x_{jk})/h\bigr)\right|\right)^2 du_i\, du_j
$$
$$
\leq \mathbb{E}\left(\frac{1}{n}\sum_{k=1}^n 2\|K\|_\infty\left(L_K\frac{\epsilon}{h}\mathbb{1}\,\Delta_k(x_i,x_j)\leq\epsilon + \mathbb{1}\,\Delta_k(x_i,x_j)\geq\epsilon\right)\right)^2
$$
$$
\leq 2C_k\left(\frac{\epsilon^2}{h^2} + e^{-2n\epsilon^2}\right) = 2C_k(\log n/(nh^2) + n^{-4}). \quad \text{(F.2)}
$$

### F.1.2  Analysis of Corner Area $\mathcal{T}_\mathcal{B}$

Recall that $\mathcal{B}$ is the four-corner area of $[0,1]^2$. In specfic, $\mathcal{B} = [0,h]^2 \cup [1-h,1]\times[0,h] \cup [0,h]\times[1-h,1] \cup [1-h,1]^2$. Without loss of generality, we analyze the left down corner $\mathcal{B}_1 = [0,h]^2$ since by symmetry, there exist a constant $C > 0$ such that

$$
\mathcal{T}_\mathcal{B} = \int_\mathcal{B} (\mathbb{E}\widetilde{c}_h(u_i, u_j) - c(u_i, u_j))^2\, du_i\, du_j \leq C\int_{\mathcal{B}_1}(\mathbb{E}\widetilde{c}_h(u_i, u_j) - c(u_i, u_j))^2\, du_i\, du_j. \quad \text{(F.3)}
$$

We denote the bivariate kernel function in terms of the rank statistics as

$$K_2^{u,h}(a,b) = K\left(\frac{u_i - a}{h}\right) K\left(\frac{u_j - b}{h}\right).$$

For $(u_i, u_j) \in \mathcal{B}_1$, according to the mirror reflect estimator defined in (4.7), we have

$$\widetilde{c}_h(u_i, u_j) = \frac{1}{nh^2} \sum_{k=1}^{n} \left[ K_2^{u,h}(\widehat{u}_{ik}, \widehat{u}_{jk}) + K_2^{u,h}(-\widehat{u}_{ik}, \widehat{u}_{jk}) + K_2^{u,h}(\widehat{u}_{ik}, -\widehat{u}_{jk}) + K_2^{u,h}(-\widehat{u}_{ik}, -\widehat{u}_{jk}) \right].$$

Similarly to the analysis of $\mathcal{T}_\mathcal{A}$, we separate it into two parts:

$$\mathcal{T}_{\mathcal{B}_1} \leq 2 \int_{\mathcal{B}_1} \left( \mathbb{E}\widetilde{c}_h - \mathbb{E}\widetilde{c}_h^{\text{std}} \right)^2 du_i\, du_j + 2 \int_{\mathcal{B}_1} \left( \mathbb{E}\widetilde{c}_h^{\text{std}} - c \right)^2 du_i\, du_j = 2(\mathcal{T}_{\mathcal{B}_1}^{(1)} + \mathcal{T}_{\mathcal{B}_1}^{(2)}).$$

Similar to the analysis in (F.1), we have $\mathcal{T}_{\mathcal{B}_1}^{(1)} \leq 2C_K(\log n\, h^{-2} n^{-1} + n^{-8})$. In order to bound $\mathcal{T}_{\mathcal{B}_1}^{(2)}$, we first study

$$\mathbb{E}\widetilde{c}_h^{\text{std}}(u_i, u_j)$$
$$= \frac{1}{h^2} \int_0^1 \int_0^1 \left[ K_2^{u,h}(s,t) + K_2^{u,h}(-s,t) + K_2^{u,h}(s,-t) + K_2^{u,h}(-s,-t) \right] c(s,t) ds dt$$
$$= \int_{-\frac{u_j}{h}}^{1} \int_{-\frac{u_i}{h}}^{1} K(s)K(t) p(u_i + sh, u_j + th) ds dt + \int_{-\frac{u_j}{h}}^{1} \int_{\frac{u_i}{h}}^{1} K(s)K(t) c(sh - u_i, u_j + th) ds dt$$
$$+ \int_{\frac{u_j}{h}}^{1} \int_{-\frac{u_i}{h}}^{1} K(s)K(t) c(sh + u_i, -u_j + th) ds dt + \int_{\frac{u_j}{h}}^{1} \int_{\frac{u_i}{h}}^{1} K(s)K(t) c(-u_i + sh, -u_j + th) ds dt.$$

By Assumption 4.4, $K(\cdot)$ is symmetric on $[-1, 1]$. We obtain

$$\int_{-\frac{u_j}{h}}^{1} \int_{\frac{u_i}{h}}^{1} K(s)K(t) ds dt = \int_{-\frac{u_j}{h}}^{1} \int_{-1}^{-\frac{u_i}{h}} K(s)K(t) ds dt,$$

$$\int_{\frac{u_j}{h}}^{1} \int_{-\frac{u_i}{h}}^{1} K(s)K(t) ds dt = \int_{-1}^{-\frac{u_j}{h}} \int_{-\frac{u_i}{h}}^{1} K(s)K(t) ds dt.$$

Hence for $(u_i, u_j) \in [0, h]^2$, we decompose the copula density function into four parts

$$c(u_i, u_j) = \int_{-\frac{u_j}{h}}^{1} \int_{-\frac{u_i}{h}}^{1} + \int_{-\frac{u_j}{h}}^{1} \int_{\frac{u_i}{h}}^{1} + \int_{\frac{u_j}{h}}^{1} \int_{-\frac{u_i}{h}}^{1} + \int_{\frac{u_j}{h}}^{1} \int_{\frac{u_i}{h}}^{1} c(u_i, u_j) K(s)K(t) ds dt.$$

Since $c \in \Sigma_\kappa(2, L)$, for $0 \leq u_i, u_j \leq h$, and $-1 \leq s, t \leq 1$, we have

$$|c(u_i + sh, u_j + th) - c(u_i, u_j)| \leq 4Lh^2, \quad |c(sh - u, u_j + th) - c(u_i, u_j)| \leq 20Lh^2,$$
$$|c(sh + u, th - u_j) - c(u_i, u_j)| \leq 20Lh^2, \quad |c(sh - u, th - u_j) - c(u_i, u_j)| \leq 36Lh^2.$$

Therefore, for any point on the left down corner $(u_i, u_j) \in \mathcal{B}_1$, we can then bound the bias term as

$$\left|\mathbb{E}\widetilde{c}_h^{\text{std}}(u_i,u_j) - c(u_i,u_j)\right| = \left|\mathbb{E}\widetilde{c}_h^{\text{std}}(u_i,u_j) - \int_{-1}^{1}\int_{-1}^{1} K(s)K(t)c(s,t)dsdt\right|$$

$$\leq \int_{-\frac{u_j}{h}}^{1}\int_{-\frac{u_i}{h}}^{1} K(s)K(t)|c(u_i+sh, u_j+th) - c(u_i,u_j)|dsdt$$

$$+ \int_{-\frac{u_j}{h}}^{1}\int_{\frac{u_i}{h}}^{1} K(s)K(t)|c(sh-u_i, u_j+th) - c(u_i,u_j)|dsdt$$

$$+ \int_{\frac{u_j}{h}}^{1}\int_{-\frac{u_i}{h}}^{1} K(s)K(t)|c(sh+u_i, th-u_j) - c(u_i,u_j)|dsdt$$

$$+ \int_{\frac{u_j}{h}}^{1}\int_{\frac{u_i}{h}}^{1} K(s)K(t)|c(sh-u_i, th-u_j) - c(u_i,u_j)|dsdt \leq 80Lh^2.$$

Combining the above inequality with the $\mathcal{T}_{\mathcal{B}_1}^{(1)}$ term and (F.3), we have the rate $T_{\mathcal{B}} \leq c[h^6 + \log n/(nh^2) + n^{-8}]$.

### F.1.3 Analysis for Margin Area $\mathcal{T}_{\mathcal{C}}$

Similar to the structure of $\mathcal{B}$, $\mathcal{C}$ has four connected components. Let $\mathcal{C}_1 = [0,h] \times [h, 1-h]$ be the left one of the four margins, similar to (F.3), there exists some constant $C > 0$ such that

$$\mathcal{T}_{\mathcal{C}} = \int_{\mathcal{C}} (\mathbb{E}\widetilde{c}_h(u_i,u_j) - c(u_i,u_j))^2 \, du_i\, du_j \leq C \int_{\mathcal{C}_1} (\mathbb{E}\widetilde{c}_h(u_i,u_j) - c(u_i,u_j))^2 \, du_i\, du_j.$$

For $(u_i, u_j) \in \mathcal{C}_1$, by the definition of mirror reflect estimator in (4.7), we have

$$\widetilde{c}_h(u_i,u_j) = \frac{1}{nh^2}\sum_{k=1}^{n}\left[K\left(\frac{u_i - \widehat{u}_{ik}}{h}\right)K\left(\frac{u_j - \widehat{u}_{jk}}{h}\right) + K\left(\frac{u_i + \widehat{u}_{ik}}{h}\right)K\left(\frac{u_j - \widehat{u}_{jk}}{h}\right)\right].$$

And we still do the partition as previous into the bias and variance:

$$\mathcal{T}_{\mathcal{C}} \leq 2\int_{\mathcal{C}}\left(\mathbb{E}\widetilde{c}_h - \mathbb{E}\widetilde{c}_h^{\text{std}}\right)^2 du_i\, du_j + 2\int_{\mathcal{C}}\left(\mathbb{E}\widetilde{c}_h^{\text{std}} - c\right)^2 du_i\, du_j =: 2(\mathcal{T}_{\mathcal{C}}^{(1)} + \mathcal{T}_{\mathcal{C}}^{(2)}).$$

Similar to the proof of (F.1), we can also bound the first term as $\mathcal{T}_{\mathcal{B}_1}^{(1)} \leq c\log n/(nh^2) + n^{-4}$.

For the second term, we plug in the estimator and calculate the expectation directly as

$$\mathbb{E}\widetilde{c}_h^{\text{std}}(u_i,u_j) = \int_0^1\int_0^1 K_h(u_i-s)K_h(u_j-t)c(s,t)dsdt + \int_0^1\int_0^1 K_h(u_i+s)K_h(u_j-t)c(s,t)dsdt$$

$$= \int_{-1}^{1}\int_{-\frac{u_i}{h}}^{1} K(s)K(t)c(u_i+sh, u_j+th)dsdt + \int_{-1}^{1}\int_{-1}^{-\frac{u_i}{h}} K(s)K(t)c(u_i-sh, u_j-th)dsdt.$$

As $c \in \Sigma_\kappa(2,L)$ for $0 < u \leq h$, by the definition of Hölder class, we have

$$|c(u_i+sh, u_j+th) - c(u_i,u_j) - \langle \nabla c(u_i,u_j), (s,t)\rangle h| \leq L(s^2+t^2)h^2,$$
$$|c(u_i-sh, u_j-th) - c(u_i,u_j) + \frac{\partial c(u_i,u_j)}{\partial u}(2u+sh) + \frac{\partial c(u_i,u_j)}{\partial u_j}(th)| \leq L[(2+s)^2+t^2]h^2.$$

For the Hölder property as $|s|,|t| \leq 1$ again, we have $|c(u_i+sh, u_j+th) - c(u_i, u_j)| \leq \left|\frac{\partial c(u_i, u_j)}{\partial u_i}\right|h + \left|\frac{\partial c(u_i, u_j)}{\partial u_j}\right|h + L(s^2+t^2)h^2$. Similarly, $|p(u_i-sh, u_j-th) - p(u_i, u_j)| \leq 9\left|\frac{\partial p(x)}{\partial u}\right|h + \left|\frac{\partial p(x)}{\partial u_j}\right|h + 10Lh^2$.

Hence we can bound the bias term on $\mathcal{C}_1$

$$\left|\mathbb{E}\widetilde{c}_h^{\text{std}}(u_i, u_j) - c(u_i, u_j)\right| = \left|\mathbb{E}\widetilde{c}_h^{\text{std}}(u_i, u_j) - \int_{-1}^{1}\int_{-1}^{1} K(s)K(t)c(s,t)dsdt\right|$$

$$\leq \int_{-1}^{1}\int_{-\frac{u_i}{h}}^{1} K(s)K(t)\left|c(u_i+sh, u_j+th) - c(u_i, u_j)\right|dsdt$$

$$+ \int_{-1}^{1}\int_{-1}^{-\frac{u_i}{h}} K(s)K(t)\left|c(-sh-u, u_j-th) - c(u_i, u_j)\right|dsdt$$

$$\leq 10\left|\frac{\partial c(u_i, u_j)}{\partial u}\right|h + 2\left|\frac{\partial c(u_i, u_j)}{\partial u_j}\right|h + 12Lh^2 \leq 12Lh^2 + 12Lh^2 = 24Lh^2,$$

where the last inequality is derived from $\left|\frac{\partial c(u_i,u_j)}{\partial u}\right|, \left|\frac{\partial c(u_i,u_j)}{\partial v}\right| \leq Lh$, by the Hölder condition and boundary assumption in Assumption 4.3. Therefore, we obtain $\mathcal{T}_\mathcal{C} \leq c[h^5 + \log n/(nh^2) + n^{-8}]$.

In summary, combining the upper bound of $\mathcal{T}_\mathcal{A}, \mathcal{T}_\mathcal{B}$ and $\mathcal{T}_\mathcal{C}$, we prove upper bound in Lemma F.1.
□

#### F.1.4 Bias of the Mutual Information Estimator

**Lemma F.2.** Under Assumption 4.3 and 4.4, we have for some constant $C_1$, the bias of mutual information estimator can be bounded by

$$\sup_{c \in \Sigma_\kappa(2,L)} \left|\mathbb{E}I\left(\widehat{c}_h\right) - I(c)\right| \leq C_1 \left(\frac{\log n}{n}\right)^{1/3},$$

*Proof.* Let $\phi(u) = u \log u$ and expand it by Taylor's Theorem, denote $\Delta c(u_i, u_j) = \widehat{c}_h(u_i, u_j) - c(u_i, u_j)$ and we get

$$\phi\left(\widehat{c}_h(u_i, u_j)\right) - \phi\left(c(u_i, u_j)\right) = \left(\log(c(u_i, u_j)) + 1\right) \cdot \Delta c(u_i, u_j) + \frac{1}{2\xi(u_i, u_j)} \cdot \Delta c^2(u_i, u_j),$$

where $\xi(u_i, u_j)$ lies in between $\widehat{c}_h(u_i, u_j)$ and $c(u_i, u_j)$. According to the boundedness of copula density $c(u_i, u_j)$ and $\widehat{c}_h$, we have $\kappa_1 \leq \xi(u_i, u_j) \leq \kappa_2$.

Let $\kappa = \max\{|\log \kappa_1|, |\log \kappa_2|\} + 1$, applying the connection between the mutual information and copula entropy in (4.6), defining $\phi(x) = x \log x$, we have

$$\left|\mathbb{E}I\left(\widehat{c}_h\right) - I(c)\right| \stackrel{(4.6)}{=} \left|\mathbb{E}H_c\left(\widehat{c}_h\right) - H_c(c)\right| = \left|\mathbb{E}\iint_{[0,1]^2}\left[\phi\left(\widehat{c}_h(u_i, u_j)\right) - \phi\left(c(u_i, u_j)\right)\right]du_i\, du_j\right|$$

$$\stackrel{\text{Fubini}}{=} \left|\iint_{[0,1]^2} \mathbb{E}\left[\phi\left(\widehat{c}_h(u_i, u_j)\right) - \phi\left(c(u_i, u_j)\right)\right]du_i\, du_j\right|$$

Applying Taylor expansion to the last equality above, we have $\left|\mathbb{E}I\left(\widehat{c}_{h}\right)-I(c)\right|\leq I_{1}+I_{2}$, where

$$I_{1}=\left|\iint_{[0,1]^{2}}\left(\log(c(u_{i},u_{j}))+1\right)\cdot\mathbb{E}[\Delta c(u_{i},u_{j})]du_{i}\,du_{j}\right|$$

$$I_{2}=\left|\iint_{[0,1]^{2}}\frac{1}{2\xi(u_{i},u_{j})}\cdot\mathbb{E}[\Delta c(u_{i},u_{j})]^{2}du_{i}\,du_{j}du_{i}\,du_{j}\right|$$

By Hölder inequality, we have $I_{1}\leq\kappa\sqrt{\iint_{[0,1]^{2}}\left[\mathbb{E}\widetilde{c}_{h}(u_{i},u_{j})-c(u_{i},u_{j})\right]^{2}du_{i}\,du_{j}}$ and

$$I_{2}\leq\frac{1}{\kappa_{1}}\iint_{[0,1]^{2}}\mathbb{E}\left[\widetilde{c}_{h}(u_{i},u_{j})-\widetilde{c}_{h}^{\mathrm{std}}(u_{i},u_{j})\right]^{2}du_{i}\,du_{j}+\frac{1}{\kappa_{1}}\iint_{[0,1]^{2}}\mathbb{E}\left[\widetilde{c}_{h}^{\mathrm{std}}(u_{i},u_{j})-c(u_{i},u_{j})\right]^{2}du_{i}\,du_{j}$$

Therefore, we can bound the bias of mutual information estimator by

$$\left|\mathbb{E}I\left(\widehat{c}_{h}\right)-I(c)\right|\leq c_{1}[h^{2}+\sqrt{\log n}/(n^{1/2}h)+n^{-4}]+c_{2}\log n/(nh^{2})+n^{-8}+c_{3}h^{4}+c_{4}h^{-2}n^{-1}.$$

The above inequality follows from Lemma F.1 and Lemma (F.2), where $c_{1},c_{2},c_{3},c_{4}$ are three constants. We prove the upper bound by setting $h\asymp(\log n/n)^{1/6}$.  $\square$

## F.2 Variance Term Analysis

We are going to show an exponential concentration inequality of our mutual information estimator.

**Lemma F.3.** *If we use the kernel function satisfies Assumption 4.4, we have*

$$\sup_{c\in\Sigma_{\kappa}(2,L)}\mathbb{P}\left(|I(\widehat{c}_{h})-\mathbb{E}H(\widehat{c}_{h})|>\epsilon\right)\leq 2\exp\left(-\frac{(n^{2}\log n)^{1/3}\epsilon^{2}}{128\kappa^{2}L_{K}^{2}}\right), \tag{F.4}$$

*where* $\kappa=\max\{|\log\kappa_{1}|,|\log\kappa_{2}|\}+1$.

*Proof.* In order to prove the concentration inequality, we apply the McDiamaid's inequality (McDiarmid, 1989). To use the inequality, we need to bound the change of the estimator if we disturb one sample point. Let $\widehat{c}'_{h}(u_{i},u_{j})$ be the kernel density estimator defined as in (4.7) by replacing $k$-th data point $(x_{ik},x_{jk})$ with an arbitrary value $((x_{ik})',(x_{jk})')$. As $\phi'(u)=\log u+1$, according to the boundedness in Assumption 4.3, we have $\max\left[|\phi'(\widehat{c}_{h}(u))|,|\phi'(\widehat{c}'_{h}(u))|\right]\leq\kappa$.

When $(x_{ik},x_{jk})$ is replaced by $((x_{ik})',(x_{jk})')$, the $(\widehat{u}_{ik},\widehat{u}_{jk})$ will correspondingly move to $((\widehat{u}_{ik})',(\widehat{u}_{jk})')$ where

$$(\widehat{u}_{im})'=\begin{cases}\widehat{u}_{im} & \text{if } \widehat{u}_{im}\notin\left(\widehat{u}_{ik}\wedge(\widehat{u}_{ik})',\widehat{u}_{ik}\vee(\widehat{u}_{ik})'\right);\\ \widehat{u}_{im}+1/n & \text{if } \widehat{u}_{ik}>(\widehat{u}_{ik})' \text{ and } \widehat{u}_{ik}\in\left[(\widehat{u}_{ik})',\widehat{u}_{ik}\right);\\ \widehat{u}_{im}-1/n & \text{if } \widehat{u}_{ik}<(\widehat{u}_{ik})' \text{ and } \widehat{u}_{ik}\in\left(\widehat{u}_{ik},(\widehat{u}_{ik})'\right],\end{cases}$$

for any $m\neq k$. Similar behaviour will happen to $(\widehat{u}_{jm})'$ as well, so we have for any $m\neq k$, $|\widehat{u}_{im}-(\widehat{u}_{im})'|\leq 1/n$ and $|\widehat{u}_{jm}-(\widehat{u}_{jm})'|\leq 1/n$.

For simplicity, we denote $K(u_i)K(u_j)$ as $K_2(u)$ and $\widehat{u}_k := (\widehat{u}_{ik}, \widehat{u}_{jk})$, $(\widehat{u}_k)' := ((\widehat{u}_{ik})', (\widehat{u}_{jk})')$. For the simplicity of notation, in the following, the supremum sup is taking over $x_{i1}, \ldots, x_{in}, (x_{ik})'$ and $x_{j1}, \ldots, x_{jn}, (x_{jk})'$. With this we have

$$\sup |I(\widehat{c}_h) - I(\widehat{c}_h')| = \sup \Big| \iint_{[0,1]^2} [\phi(\widehat{c}_h(u_i, u_j)) - \phi(\widehat{c}_h'(u_i, u_j))] du_i \, du_j \Big|$$

$$\leq \kappa \sup \iint_{[0,1]^2} |\widehat{c}_h(u_i, u_j) - \widehat{c}_h'(u_i, u_j)| du_i \, du_j \leq \kappa \sup \iint_{[0,1]^2} |\widetilde{c}_h(u_i, u_j) - \widetilde{c}_h'(u_i, u_j)| du_i \, du_j.$$

By the definition of the kernel density estimator, we have

$$\sup |I(\widehat{c}_h) - I(\widehat{c}_h')| \leq 4\kappa \sup \bigg( \iint_{[0,1]^2} \Big| \frac{1}{nh^2} \sum_{m \neq k} K_2\Big(\frac{u - \widehat{u}_m}{h}\Big) - \frac{1}{nh^2} \sum_{m \neq k} K_2\Big(\frac{u - (\widehat{u}_m)'}{h}\Big) \Big| du$$

$$+ \iint_{[0,1]^2} \Big| \frac{1}{nh^2} K_2\Big(\frac{u - \widehat{u}_k}{h}\Big) - \frac{1}{nh^2} K_2\Big(\frac{u - (\widehat{u}_k)'}{h}\Big) \Big| du \bigg)$$

$$\leq \frac{4\kappa}{n} \sum_{m \neq k} \iint_{[0,1]^2} \Big| K_2\Big(u - \frac{\widehat{u}_k}{h}\Big) - K_2\Big(u - \frac{(\widehat{u}_k)'}{h}\Big) \Big| du$$

$$+ 8\kappa \sup_{s \in \mathbb{R}^2} \iint_{[0,1]^2} \frac{1}{nh^2} K_2\Big(\frac{s - u}{h}\Big) du,$$

where the last inequality is due to the change of variable. By the Lipschitz property of kernel, we can further bound the difference by

$$\sup |I(\widehat{c}_h) - I(\widehat{c}_h')| \leq \frac{4\kappa}{n} \sum_{m \neq k} \frac{L_K}{h} ||\widehat{u}_m - (\widehat{u}_m)'||_2^2 + \frac{8\kappa}{n} \leq \frac{8\kappa L_K}{nh} + \frac{8\kappa}{n} \leq \frac{16\kappa L_K}{nh}.$$

Using McDiamaid's inequality we have

$$\sup_{c \in \Sigma_\kappa(2, L)} \mathbb{P}\left( |I(\widehat{c}_h) - \mathbb{E} H(\widehat{c}_h)| > \epsilon \right) \leq 2 \exp\left( -\frac{nh^2 \epsilon^2}{128 \kappa^2 L_K^2} \right),$$

and we prove the exponential concentration inequality (F.4) by setting $h \asymp (\log n / n)^{1/6}$ same as the bandwidth in Lemma F.2. $\square$

## G  Proof of Theorem 4.10

*Proof.* We define the $L_1$-norm $\|p\|_1 := \int_{\mathcal{X}} |p(x)| dx$ and

$$\widehat{p}_{F^*}(x) = \prod_{(j,k) \in E^*} \frac{\widetilde{p}_{h_2}(x_j, x_k)}{\widetilde{p}_{h_2}(x_j) \widetilde{p}_{h_2}(x_k)} \cdot \prod_{u \in U^*} \widetilde{p}_{h_2}(x_u) \cdot \prod_{\ell \in V \setminus U^*} \widetilde{p}_{h_1}(x_\ell),$$

where $U^* \subset V$ is the set of isolated vertices in the oracle-forest $F^*$.

It is easy to see that $\widehat{p}_{F^*}(x) \geq 0$ and $\int \widehat{p}_{F^*}(x) dx = 1$. Let $D(p\|q)$ be the KL-divergence between two densities $p$ and $q$. Using Pinsker's inequality, we have

$$\|\widehat{p}_{F^*} - p_{F^*}\|_1 \leq \min\{\sqrt{D(p_{F^*}\|\widehat{p}_{F^*})}, \sqrt{D(\widehat{p}_{F^*}\|p_{F^*})}\}, \text{ where} \tag{G.1}$$

$$D(p_{F^*}\|\widehat{p}_{F^*}) = \int p_{F^*}(x) \log \frac{p_{F^*}(x)}{\widehat{p}_{F^*}(x)} dx$$

$$= \int p_{F^*}(x) \log \frac{\prod_{(j,k)\in E^*} p(x_j, x_k)}{\prod_{(j,k)\in E^*} \widehat{p}_{h_2}(x_j, x_k)} dx - \int p_{F^*}(x) \log \frac{\prod_{(j,k)\in E^*} p(x_j)p(x_k)}{\prod_{(j,k)\in E^*} \widehat{p}_{h_2}(x_j)\widehat{p}_{h_2}(x_k)} dx$$

$$+ \sum_{u\in U^*} \int p(x_u) \log \frac{p(x_u)}{\widehat{p}_{h_2}(x_u)} dx_u + \sum_{\ell\in V\setminus U^*} \int p(x_\ell) \log \frac{p(x_\ell)}{\widehat{p}_{h_1}(x_\ell)} dx_\ell.$$

Therefore, the expectation of the KL-divergence can be bounded by

$$\mathbb{E} D(p_{F^*}\|\widehat{p}_{F^*}) \leq s \cdot \max_{jk} \mathbb{E} \int p(x_j, x_k) \log \frac{p(x_j, x_k)}{\widehat{p}_{h_2}(x_j, x_k)} dx_j dx_k + s \cdot \max_u \mathbb{E} \int p(x_u) \log \frac{p(x_u)}{\widehat{p}_{h_2}(x_u)} dx_u$$

$$+ (d-s) \cdot \max_\ell \mathbb{E} \int p(x_\ell) \log \frac{p(x_\ell)}{\widehat{p}_{h_1}(x_\ell)} dx_\ell$$

$$= s \cdot \max_{jk} \mathbb{E} D(p_{jk}\|\widehat{p}_{jk}) + s \cdot \max_u \mathbb{E} D(p_u\|\widehat{p}_u^{(2)}) + (d-s) \cdot \max_\ell \mathbb{E} D(p_\ell\|\widehat{p}_\ell^{(1)}),$$

where $\widehat{p}_{jk} := \widehat{p}_{h_2}(x_j, x_k), \widehat{p}_u^{(2)} := \widehat{p}_{h_2}(x_u), \widehat{p}_\ell^{(1)} := \widehat{p}_{h_1}(x_\ell)$. Similarly, we have

$$\mathbb{E} D(\widehat{p}_{F^*}\|p_{F^*}) \leq s \cdot \max_{jk} \mathbb{E} D(\widehat{p}_{jk}\|p_{jk}) + s \cdot \max_u \mathbb{E} D(\widehat{p}_u^{(2)}\|p_u) + (d-s) \cdot \max_\ell \mathbb{E} D(\widehat{p}_\ell^{(1)}\|p_\ell). \quad \text{(G.2)}$$

Since $\kappa_1 \leq \inf_{jk} p_{jk}$, the KL-divergence can be further bounded by

$$D(\widehat{p}_{jk}\|p_{jk}) = \int \widehat{p}_{h_2}(x_j, x_k) \log \frac{\widehat{p}_{h_2}(x_j, x_k)}{p(x_j, x_k)} dx_j dx_k \leq \int \frac{(\widehat{p}_{h_2}(x_j, x_k))^2}{p(x_j, x_k)} dx_j dx_k - 1$$

$$= \int \frac{(\widehat{p}_{h_2}(x_j, x_k) - p(x_j, x_k))^2}{p(x_j, x_k)} dx_j dx_k \leq \frac{1}{\kappa_1} \|p_{jk} - \widehat{p}_{jk}\|_2^2. \quad \text{(G.3)}$$

Using (G.3), we know that there exists a constant $C_1$ which does not depend on $j, k$, such that

$$\sup_{p_{jk}\in\Sigma_\kappa(2,L)} \mathbb{E} D(\widehat{p}_{jk}\|p_{jk}) \leq \frac{1}{\kappa_1} \sup_{p_{jk}\in\Sigma_\kappa(2,L)} \mathbb{E}\|p_{jk} - \widehat{p}_{jk}\|_2^2 \leq C_1 n^{-\frac{2}{3}}.$$

Similarly, we control the terms $\mathbb{E} D(\widehat{p}_u^{(2)}\|p_u)$ and $\max_\ell D(\widehat{p}_\ell^{(1)}\|p_\ell)$ and show that they have the rate $C_2 \cdot n^{-2/3}$ and $C_3 \cdot n^{-4/5}$ respectively.

Therefore, by (G.1) and (G.2) we have for some constant $C$,

$$\sup_{F\in\mathcal{F}_d^s} \sup_{p\in\Sigma_\kappa(2,L)} \mathbb{E}\|\widehat{p}_{F^*} - p_{F^*}\|_1 \leq C \cdot \sqrt{\frac{s}{n^{2/3}} + \frac{d-s}{n^{4/5}}},$$

Using the fact that $\|\widehat{p}_{\widehat{F}} - p_{F^*}\|_1 \leq \|\widehat{p}_{\widehat{F}}\|_1 + \|p_{F^*}\|_1 \leq 2$, we have

$$\mathbb{E}\|\widehat{p}_{\widehat{F}} - p_{F^*}\|_1 \leq \mathbb{E}\|\widehat{p}_{F^*} - p_{F^*}\|_1 + \mathbb{E}\|\widehat{p}_{\widehat{F}} - p_{F^*}\|_1 \cdot I(\widehat{F} \neq F^*)$$

$$\leq C \cdot \sqrt{\frac{s}{n^{2/3}} + \frac{d-s}{n^{4/5}}} + 2\mathbb{P}\left(\widehat{F} \neq F^*\right).$$

From (D.1) in the proof of Theorem 3.6, we have

$$\sup_{p \in \mathcal{P}_\kappa} \mathbb{P}\left(\widehat{F} \neq F^*\right) = O\left(d^{-(\log n)^{1/3}}\right).$$

The desired result complete the proof of theorem. □